\documentclass[pdflatex,sn-mathphys-ay]{sn-jnl}

\usepackage{graphicx}%
\usepackage{multirow}%
\usepackage{amsmath,amssymb,amsfonts}%
\usepackage{mathrsfs}%
 \usepackage{mathtools}%
\usepackage[title]{appendix}%
\usepackage{xcolor}%
\usepackage{textcomp}%
\usepackage{manyfoot}%
\usepackage{booktabs}%
\usepackage{algorithm}%
\usepackage{algorithmicx}%
\usepackage{algpseudocode}%
\usepackage{makecell}  
\usepackage{array}
\usepackage{float}
\usepackage{tabularx}
\usepackage{longtable}
\usepackage{subcaption}
\usepackage{natbib}
\usepackage[printonlyused,nohyperlinks]{acronym}
\usepackage{units}
\usepackage[utf8]{inputenc}
\usepackage{linguex}
\usepackage{csquotes}
\usepackage{dirtytalk}
\usepackage{supertabular}
\usepackage[section]{placeins}
\usepackage{adjustbox}
\usepackage{comment}
\usepackage{hyperref}
\usepackage{qtree}
\usepackage{ltablex}
\keepXColumns
\usepackage[left]{lineno}

\newcommand{\modelEL}{\textsc{ELD}}

\newcommand{\modelCBLINK}{\textsc{C-BLINK}}

\newcommand{\benchmarkEL}{\textsc{MHERCL}}

\usepackage[inline]{enumitem}
\usepackage{adjustbox}
\usepackage{caption}
\usepackage{subcaption}
\usepackage{listings}
\lstdefinelanguage{SPARQL}{    morekeywords={SELECT,WHERE,FILTER,OPTIONAL,GRAPH,PREFIX,CONSTRUCT,ASK,DESCRIBE,ORDER,BY,ASC,DESC,LIMIT,OFFSET,FROM,NAMED,GROUP,HAVING,UNION,AND,NOT,EXISTS,MINUS,BIND,VALUES,SERVICE,SILENT,IN,LOAD,CLEAR,CREATE,DROP,COPY,MOVE,ADD,TO,AS,WITH,ALL,DATA,INTO},
    sensitive=true,
    morecomment=[l]{\#},
    morestring=[b]",
}
\lstdefinelanguage{RDF}{
    morekeywords={@prefix, a},
    sensitive=false,
    morecomment=[l]{\#},
    morestring=[b]",
}
\lstdefinelanguage{ConLL}{
}
\newtheorem{example}{Example}

\makeatletter 
\def\@acknow{}%
\long\def\EarlyAcknow#1 \par{%
\def\@acknow{\abstractfont\subabstracthead*{Acknowledgments}
#1\par}}%

\def\printabstract{\ifx\@acknow\empty\else\@acknow\fi\par%
    \ifx\@abstract\empty\else\@abstract\fi\par}
\makeatother

\EarlyAcknow{This project has received funding from the European Union’s Horizon 2020 research and innovation programme
under grant agreement No 101004746. The authors acknowledge Dr Enrico Daga's feedback on this manuscript, which provided an insightful external perspective.
}
\vspace{-5mm}  

\begin{document}

\title[Article Title]{Musical Heritage Historical Entity Linking}
\author*[1]{\fnm{Arianna} \sur{Graciotti}}\email{arianna.graciotti@unibo.it}
\author[1,2]{\fnm{Nicolas} \sur{Lazzari}}\email{nicolas.lazzari3@unibo.it}
\author[1]{\fnm{Valentina} \sur{Presutti}}\email{valentina.presutti@unibo.it}
\author[3]{\fnm{Rocco} \sur{Tripodi}}\email{rocco.tripodi@unive.it}
\equalcont{Authors are listed in alphabetical order.}

\affil*[1]{\orgdiv{LILEC}, \orgname{University of Bologna}, \orgaddress{\country{Italy}}}
\affil[2]{\orgdiv{Computer Science Department}, \orgname{University of Pisa}, \country{Italy}}
\affil[3]{\orgname{DAIS, Ca' Foscari University of Venice}, \country{Italy}}

\abstract{Linking named entities occurring in text to their corresponding entity in a Knowledge Base (KB) is challenging, especially when dealing with historical texts. In this work, we introduce Musical Heritage named Entities Recognition, Classification and Linking (\benchmarkEL{}), a novel benchmark consisting of manually annotated sentences extrapolated from historical periodicals of the music domain. \benchmarkEL{} contains named entities under-represented or absent in the most famous KBs. We experiment with several State-of-the-Art models on the Entity Linking (EL) task and show that \benchmarkEL{} is a challenging dataset for all of them. We propose a novel unsupervised EL model and a method to extend supervised entity linkers by using Knowledge Graphs (KGs) to tackle the main difficulties posed by historical documents. Our experiments reveal that relying on unsupervised techniques and improving models with logical constraints based on KGs and heuristics to predict \texttt{NIL} entities (entities not represented in the KB of reference) results in better EL performance on historical documents.}


\maketitle

\section{Introduction}

\label{sec:hel-introduction}
Named Entity Recognition (NER), Named Entity Classification (NEC), and Entity Linking (EL) -- also referred to as Entity Disambiguation \citep{bunescu-pasca-2006-using} --  are essential tasks in Knowledge Extraction (KE). NER involves identifying spans of text (\textit{mentions}) that refer to named entities, such as people, locations, or organizations. NEC then assigns a specific entity type to each mention according to a predefined set of categories. EL links each mention to a unique entity within a reference Knowledge Base (KB), such as Wikipedia, DBPedia, or Wikidata. This linking process considers the context in which the mention occurs and, in some cases, information from the NER and NEC tasks \citep{tedeschi-etal-2021-named-entity}.

The extensive digitization of historical documents\footnote{We define historical documents as any textual materials produced or published up to 1979, as done by \citet{ehrmannNamedEntityRecognition2023}.} is fundamental to the study and preservation of cultural heritage. The automatic processing of these documents revealed new challenges for NER, NEC, and EL models, deriving from the noisy quality of digitized historical text resulting from the use of Optical Character Recognition (OCR) technology or from variations between contemporary and historical language, such as differences in spelling, sentence structure, or changes in entities names over time.

\begin{figure}[t]
    \centering
    \includegraphics[width=\linewidth]{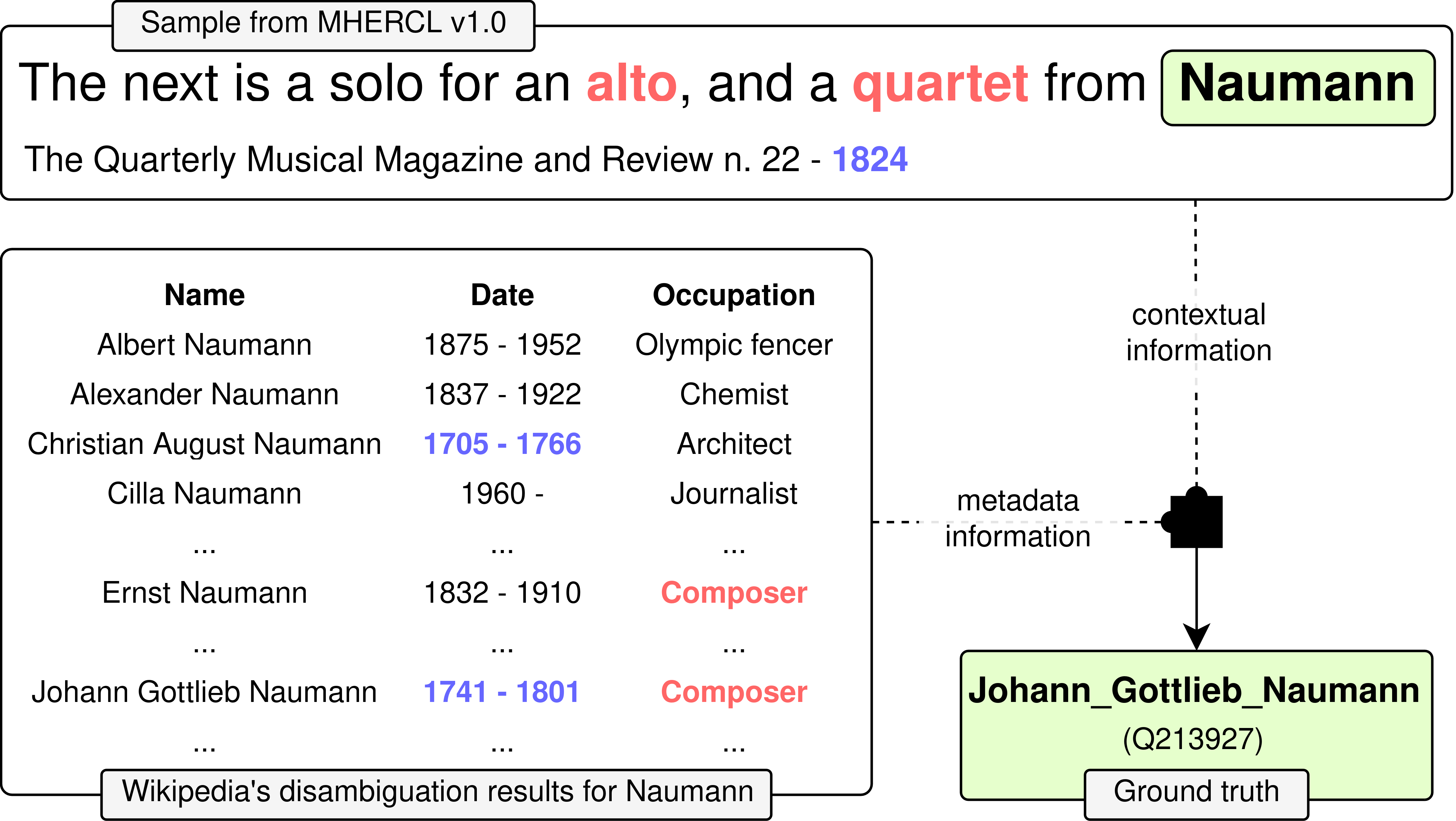}
    \caption{Example sentence from an 1824 document in the Polifonia Corpus, with \textit{Naumann} as the entity to be linked. The figure includes a selection of results from the Wikipedia disambiguation page (\url{https://en.wikipedia.org/wiki/Naumann}). Using sentence context and metadata from both the entity and the document allows for identifying the plausible entity, which in this case is unique.}
    \label{fig:plausibility}
\end{figure}

In addition to these challenges, the scarcity of annotated resources poses a significant barrier to Historical Entity Linking (HEL). State-of-the-art (SotA) EL models, trained on contemporary web-based datasets, perform poorly on historical texts, where entities occur infrequently or are absent in the training data \citep{hoffart-etal-2011-robust, chen-etal-2021-evaluating, mollerSurveyEnglishEntity2022}. For instance, consider the sentence ``The next is a solo for an alto, and a quartet from \underline{Naumann}'' in Figure \ref{fig:plausibility}, extracted from the 1824 issue of the historical music periodical \textit{The Quarterly Musical Magazine And Review}, part of the Polifonia Corpus\footnote{The Polifonia Corpus is a multilingual and diachronic corpus focused on Musical Heritage. Available at \url{https://github.com/polifonia-project/Polifonia-Corpus}.}\label{foot:polifonia_corpus}. The mention of \textit{Naumann} is challenging to resolve: the Wikipedia disambiguation page lists over 30 possible entities\footnote{Available at \url{https://en.wikipedia.org/wiki/Naumann}, last accessed May 22, 2024.}, each appearing infrequently (0 to 3 times) in typical training sets \citep{wu-etal-2020-scalable}. Understanding the context of the mention aids in linking it to the correct entity. Music-specific terms in the sentence suggests a person relevant to the music domain. Knowing the date of the source document allows additional filtering, as entities born after 1824 shall be excluded. SotA models that ignore such contextual and temporal cues struggle with HEL, which is inherently more complex than linking entities in contemporary texts.

Furthermore, entity linkers tend to exhibit \textit{popularity bias} \citep{kandpalLargeLanguageModels2023}, favouring frequently occurring entities in their training sets over less common (\textit{long-tail}) ones. This bias is particularly problematic in historical documents, where entities are often less prominent than contemporary ones and appear less frequently in widely used KBs such as Wikipedia.


Finally, historical documents often contain named entity mentions without corresponding entities in a KB, known as \texttt{NIL} links \citep{sevgiliNeuralEntityLinking2022, guellilEntityLinkingEnglish2024c}. Even though explicitly handling \texttt{NIL} links is fundamental in HEL, most SotA entity linkers lack this capability.

Following these arguments, our contribution is three-fold:

\begin{enumerate}
\item We release Musical Heritage Historical named Entities Recognition, Classification and Linking (\benchmarkEL{}), a gold-standard resource of historical documents focused on musical heritage and annotated for NER, NEC and EL. Although domain-specific, \benchmarkEL{} aligns with typical HEL resources, which are mainly derived from newspapers \citep{HIPE2022ExtOverview} and is composed of a high number of long-tail entities with respect to their popularity.
\item We propose Entity Linking Dynamics (\modelEL{}), an unsupervised EL model based on a game-theoretical approach where mentions and entities interact to reach a consensus on the correct link. This approach addresses popularity bias by avoiding a supervised training phase, enabling the model to perform well on low-frequency entities often present in historical texts.
By using an unsupervised approach, we bypass the limitations of labelled data, which may not cover all possible historical variations and interpretations of an entity, leading to potentially stronger generalization in scenarios where labelled data is sparse or incomplete.
\item We introduce a method that leverages constraints extracted from a Knowledge Graph (KG) to reduce implausible linking in retrieval-based EL models. We extend \citep[BLINK]{wu-etal-2020-scalable} and propose Constrained-BLINK (\modelCBLINK{}), a version that incorporates these constraints, allowing for improved performance on historical texts. Moreover, we systematically explore how \modelCBLINK{} can be extended to account for \texttt{NIL} links, which are widely found in historical documents.
\end{enumerate}

The code for our experiments is publicly available on GitHub\footnote{Available at the URL \url{https://github.com/polifonia-project/historical-entity-linking}.}\label{foot:github}, and the dataset is also available on HuggingFace\footnote{Available at \url{https://huggingface.co/datasets/n28div/MHERCL}.}\label{foot:huggingface}.

This paper is organised as follows: Section \ref{sec:related} reviews key contributions on EL benchmarks and models, focusing on those addressing the specific challenges of historical documents. Section \ref{sec:mhercl} describes the documents selected for creating \benchmarkEL{} and provides details on annotation guidelines and dataset statistics. In Section \ref{sec:models}, we outline the implementation of \modelEL{} and describe how we adapted BLINK with typological and temporal constraints to create \modelCBLINK{}. Section \ref{sec:experiments} presents experimental results. Finally, Section \ref{sec:conclusion} summarizes our contributions and suggests directions for future research.

\section{Related works} 
\label{sec:related}
In this section, we examine relevant datasets and benchmarks (Section \ref{sec:related:bench}) as well as models (Section \ref{sec:related:models}) pertaining to EL. We present a detailed analysis of both general-purpose and historical document-specific benchmarks and methodologies.

\subsection{Benchmarks} \label{sec:related:bench}
\subsubsection*{General purpose}
Recent surveys indicate that historical texts are rarely utilized in datasets for training, fine-tuning, or evaluating EL models \citep{mollerSurveyEnglishEntity2022, guellilEntityLinkingEnglish2024c}. The most commonly used EL datasets for the English language include AIDA-CONLL \citep{hoffart-etal-2011-robust}, which contains Reuters news stories from 1996-1997; AQUAINT \citep{Graff2002Aquaint}, a derivative of AIDA-CONLL; MSNBC (English), which is made of online news from 2007; ACE2004 \citep{Mitchell2005}, which is composed of web-collected news articles (also in Chinese and Arabic, other than in English); CWEB \citep{Gabrilovich2013}, which consists of Wikipedia and generic web pages; WIKI \citep{10.3233/SW-170273}, which is made of Wikipedia pages; TAC \citep{TAC2010}, which contains Wikipedia pages extracted in 2008; and KORE50 \citep{KORE2012}, which is composed of web documents with a focus on highly ambiguous entities.

Several studies have targeted particularly challenging EL scenarios. The ShadowLink dataset \citep{provatorova-etal-2021-robustness} includes 16,000 short text snippets from websites in English, annotated with highly ambiguous entities. This dataset focuses on named entities with identical surface forms linked to common and rare entities, causing the former to overshadow the latter. The TempEL dataset \citep{zaporojets2022tempel} offers time-stratified English Wikipedia snapshots from 2013 to 2022, enabling the evaluation of EL models on both continuous and newly emerging entities. Additionally, \citet{benkhedda-etal-2024-enriching} concentrated on the English language Community-generated digital content (CGDC) for digital archives and developed a dataset of textual metadata linked to Wikidata information. Although this metadata includes historical named entities due to its intended use (enhancing documentation for historical digital archives), it remains a product of contemporary digital text and does not exhibit the distinctive features of historical documents.

\citet{zhu-etal-2023-learn} focuses on entity mentions that cannot be resolved against the reference KB and introduces NEL, an annotated dataset of English Wikipedia excerpts with a significant percentage of \texttt{NIL}, i.e. entities that do not exist in the KB of reference. The significant presence of \texttt{NIL} entities is particularly relevant also for HEL, as historical documents present a high occurrence of out-of-KB entities. Despite addressing \texttt{NIL} links, this work concentrated on digitally-born texts, which display different linguistic characteristics compared to digitalized documents.

\subsubsection*{Historical}
\citet{ehrmannNamedEntityRecognition2023} reviews resources for historical NER. Of particular relevance for the work described in this paper are the resources collected for the HIPE-2020 \citep{HIPE2020Overview,HIPE2020ExtOverview} and HIPE-2022 \citep{HIPE2022Overview} evaluation campaigns, as they include EL annotations.
In particular, the HIPE-2022 corpus comprises six datasets spanning English, Finnish, French, German, and Swedish. The datasets are composed by collecting historical newspapers and classic commentaries of 200 years time-span 
sourced from:
\begin{enumerate*}[label=(\roman*)]
    \item \emph{NewsEye} \citep{Hamdi2021}, which includes historical newspaper articles in French, German, Finnish, and Swedish dating back to 19C-20C;
    \item \emph{SoNAR} \citep{Sonar2021}, which includes historical newspaper articles from the Berlin State Library newspaper collections in German (19C-20C);
    \item \emph{Le Temps} \citep{ehrmann2016diachronic}, which contains historical newspaper articles from two Swiss newspapers in French (19C-20C);
    \item \emph{Living with Machines} \citep{ehrmann2016diachronic}, which contains \textit{TopRes19th}, a dataset of historical newspaper articles from the British Library newspapers in English (18C-19C), whose annotations focus on toponyms documents;
    \item \emph{AjMC}
    , which contains annotated \textit{Classical Commentaries} (19C)
\end{enumerate*}.
We employ the HIPE-2020 English test set in our experiments, as detailed in Section \ref{sec:experiments}.

Another effort dedicated to historical documents brought to the creation of Giorgio Vasari's \textit{Lives of The Artists} \citep{santini2022vasari}, which contains NER, NEC and EL annotations towards Wikidata.
A similar initiative is followed by \citet{blouin-etal-2024-dataset-named}, which carries out NER and EL towards Wikidata annotations on a historical newspaper in the Chinese language published between 1872 and 1949 and on bilingual (Chinese-English) biographies dating back to the first half of the 20C.

Our resource enriches the field of HEL, which suffers from a scarcity of resources. It addresses the music domain, a sector not yet covered by existing HEL resources. Moreover, while contemporary EL benchmarks often emphasize English, HEL resources are not predominantly in English. For example, our dataset contains almost five times as many annotated named entities as the HIPE-2020 English test set, positioning itself as a valuable complement to existing resources for NER, NEC and EL on historical documents in English.
 
\subsection{Models} \label{sec:related:models}
\subsubsection*{General purpose}
Current research on EL models predominantly employs Pre-trained Language Models (PLM). Two primary approaches have been identified: retrieval-based and generative methods. Retrieval-based methods involve encoding mentions into a dense representation, which is then compared with a similarly dense representation calculated from the documents in a KB, such as Wikipedia. An optional re-ranking phase can merge these dense representations for improved accuracy. This approach is exemplified by tools like \citep[REL]{REL2020}, which features a modular architecture of neural components, as well as BLINK and its multilingual extension, \citep[BELA]{plekhanov2023multilingual}.

A similar approach is proposed in \citep[EntQA]{zhang2022entqa} and \citep[ReLiK]{Orlando2024}. Unlike in BLINK, the linking phase is framed as a Question Answering (QA) problem.
A plausible set of entities is first retrieved using a dense representation similar to the one from BLINK.
The difference is that the named entity is not recognized beforehand, but the model jointly recognizes mentions and possible entities that might be linked.
The correct entity is then chosen based on the context of the sentence.

\citep[CLOCQ]{Christmann2022clocq} is a system designed to extend beyond the task of linking named entities to their corresponding entries in a knowledge base, addressing the broader challenge of complex question answering over knowledge bases. One of its core functionalities includes disambiguating named entities, making it relevant to our work.

A different approach, based on generative language models instead of classification architectures, is proposed in \citep[GENRE]{decao2021autoregressive}.
This approach retrieves entities by generating their unique names, left to right, conditioned on the context.
Such an approach directly captures relations between context and entity name and reduces the memory footprint of performing queries on large search spaces.
An extension of this model, with support for multiple languages, is proposed in \citep[mGENRE]{de-cao-etal-2022-multilingual}. 
\citep[ExtEnD]{barba-etal-2022-extend} builds upon the approach used by mGENRE, but reformulates it by framing that as a text extraction task.
The set of generated candidates is concatenated to the span, thereby overcoming GENRE's limitation of being unable to see the candidate entities from which to choose.

EL models that leverage symbolic knowledge from KBs can significantly enhance the plausibility of linking entities.
As demonstrated by \citet{tedeschi-etal-2021-named-entity}, it is possible to outperform traditional methods and minimize confusion during the linking process by filtering the pool of linkable entities based on the properties of the mention.
This method involves classifying named entities mentions into fine-grained categories, ensuring that potentially linkable entities share the same type as the mention.
Consequently, the model focuses on determining the most likely entity rather than learning the plausibility of each entity implicitly.

Building on similar assumptions, \citep[ReFinED]{ayoola-etal-2022-improving} focuses on optimizing performances on long-tail entities by leveraging symbolic knowledge extrapolated from KBs (entities descriptions', types, and inter-entities relations in context). Given its focus on long-tail entities, ReFinED encompasses the management of out-of-KB entities by including the \texttt{NIL} label as a possible candidate for the linking step.

Except for ReFiNeD, SoTA entity linkers presuppose that a correct link for each mention is always present in the KB of reference.
As a result, impaired focus has been dedicated by the research community to the \texttt{NIL} prediction issue.
Different approaches have been proposed to predict whether a \texttt{NIL} prediction should be inferred from the similarity score.
This includes heuristic methods, such as a threshold on the score, or machine-learning-based methods, such as training a binary classifier \citep{sevgiliNeuralEntityLinking2022}. To address this issue, \citet{10.1145/3583780.3615036} propose BLINKout, a method that uses a dynamic \texttt{NIL} representation and classification approach, incorporating synonyms and built on BERT-based EL.
 
\subsubsection*{Historical}
The scarcity of established resources and benchmarks for HEL results in fewer solutions than conventional EL.
\citet{agarwal-etal-2018-dianed} pioneered the research in this field by proposing diaNED, a time-aware entity disambiguation approach for diachronic corpora, which leverages KB-driven temporal information and text-driven temporal expressions.

The HIPE-2022 evaluation campaign collates the most recent efforts in implementing models for NER, NEC and EL on historical documents.
The highest scoring model for the challenge on CLEF-HIPE-2022 English language data for EL was proposed by the L3i team \citep{borosKnowledgebasedContextsHistorical}, which builds upon the same multilingual model based on a BiLSTM architecture \citep{kolitsas-etal-2018-end} that the team presented when they introduced it for CLEF-HIPE-2020 \citep{boros2020robust}. 
Their model is enhanced with a filtering process \citep{linharespontes2022melhissa} to disambiguate historical references using the typological and temporal information in Wikidata. All the models that took part in the evaluation campaign implemented strategies to take into consideration the prediction of the \texttt{NIL} link.

In this paper, we introduce \modelCBLINK{}, which enriches the scenario of the aforementioned related works by extending BLINK.
Our model leverages time- and type-related knowledge from our chosen KB of reference, Wikidata, to refine the candidate generation process.
After filtering out the candidates that fail the plausibility criteria scoped by the time- and type-related constraints, we test various methods, including threshold-based and machine learning-based approaches, to identify \texttt{NIL} links. The original work does not cover both aspects.

Also, we propose \modelEL{}, which differs from existing approaches in HEL by framing the El problem using a game-theoretical approach. Named entity mentions, their surrounding context, and KB entities are embedded into dense vectors. The resulting vectors are used to compute pairwise similarities. By playing against each other, named entities have to collaborate with most similar entities, eventually converging to the final link decision. The model is unsupervised.

\section{MHERCL: a new benchmark for HEL} 
\label{sec:mhercl}
This section describes \benchmarkEL{}, the new benchmark we contribute for NER, NEC and EL on historical documents.
We describe how we build the sample of sentences, the annotation guidelines, and the Inter-Annotator Agreement (IAA)  alongside the benchmark statistics.

\subsection{Sampling} \label{sec:sampling}
\benchmarkEL{} contains manually annotated sentences extrapolated from the Polifonia Corpus\footref{foot:polifonia_corpus}.
The Polifonia Corpus is a diachronic and multilingual corpus specialized in the music domain.
It is composed of multiple modules, which contain documents from various data sources.
A subset of the \textit{Periodicals} module, containing documents whose publication dates range from 1823 to 1900, is used as the set of sentences within \benchmarkEL{}.
The subset is selected such that sentences are in English and contain at least one named entity.
Because the Polifonia Corpus's focus is on music, it provides a unique domain-specific perspective on the EL problem. Nonetheless, the documents within \benchmarkEL{} align closely with the prevailing document types already utilized in existing benchmarks within the field, namely historical newspapers \citep{HIPE2022ExtOverview}. 

\subsection{Data Annotation} 
\label{subsub:ann_guid}
The sentences included within the selected subset are manually annotated by two annotators selected from Foreign Languages and Literature undergraduate students at the University of Bologna.
Both annotators have been trained on the NER, NEC and EL annotation tasks.

The annotation work was performed under the criteria that a named entity is a real-world thing indicating a unique individual through a proper noun \citep{JurafskyMartin2023}.
The annotation process involves inspecting the sentences and identifying the named entities, eventually linking them to their corresponding Wikidata unique identifier (QID).

Named entities are recognized, classified, and linked following the Abstract Meaning Representation (AMR, \citep{Banarescu2013}) named entity annotations guidelines.\footnote{Available at the URL \url{https://github.com/amrisi/amr-guidelines/blob/master/amr.md\#named-entities}, last time accessed on August 12th, 2024.\label{foot:amrguidelines}} The guidelines report instructions on how to classify the named entities according to a pre-defined list of types.

A full description of the types used for NEC in \benchmarkEL{} is provided in Appendix \ref{app:A}.
Annotation guidelines are extensively described in Appendix \ref{app:B}.

\subsection{Inter-annotator agreement} \label{subsub:IAA}
Inter-annotator agreement (IAA) measures the reliability of human annotations by assessing the consistency among annotators.
To evaluate IAA, two annotators independently annotate the same 101 sentences extrapolated from \benchmarkEL{}.
We compute Krippendorff's alpha \citep{Krippendorff2007, castro-2017-fast-krippendorff} for nominal metrics on the resulting annotations.
We choose Krippendorff's alpha due to its flexibility and resilience in handling missing values, as demonstrated in other contexts \citep{tripodi-etal-2022-evaluating}.
Table \ref{tab:MHERCLv01_filtered_IAA_Stats} presents the number of sentences and tokens of the IAA sample.

\begin{table}[ht]
    \caption{IAA within \benchmarkEL.}
    \centering
    \begin{tabular}{ccccccc}
        \toprule
        & & \multicolumn{2}{c}{\#tokens (Annotated)} & \\
        \#sents & \#tokens (Tot.) & \#matching & \#unmatching & Krippendorf's alpha \\ \midrule
        101 & 6,589 & 656 & 124 & 0.82 \\
        \bottomrule
    \end{tabular}
    
    \label{tab:MHERCLv01_filtered_IAA_Stats}
\end{table}

With an annotator agreement higher than $0.8$, \benchmarkEL{}'s annotation quality is reliable \citep{Krippendorff2007} and can be treated as a gold standard.

When we calculate the IAA, we identify instances where the annotations of the two annotators do not match, as shown in Table \ref{tab:MHERCLv01_filtered_IAA_Stats}.
We observe that this discrepancy is often due to the complexity of the HEL tasks.
For example, historical names used when the periodicals were issued might not be used anymore in current times.

Consider the following example sentence from \benchmarkEL{}, extrapolated from an issue of the music periodical \textit{The Harmonicon}, dating back to 1833:
\begin{example}\label{ex:court-magazine}
\textit{I find the following sensible critical remarks among my papers; they were copied from a weekly work (the Court Magazine, I think)}
\end{example}

In Example \ref{ex:court-magazine}, one annotator linked the mention ``Court Magazine'' to ``La Belle Assemblée'' \footnote{Available at the URLs \url{https://en.wikipedia.org/wiki/La_Belle_Assemblée} and \url{https://www.Wikidata.org/wiki/Q3206551}, last time accessed on August 12th, 2024.} while the other annotator labeled it as \texttt{NIL}. As reported in Wikipedia, ``Court Magazine'' resulted from the historical merging of different periodicals, including ``La Belle Assemblée'':

\begin{quote}
\textit{The magazine was published as \textit{La Belle Assemblée} from 1806 until May 1832. It became \textit{The Court Magazine and Belle Assemblée} from 1832 to 1837. After 1837, the \textit{Belle Assemblée} name was dropped when the magazine merged with the \textit{Lady's Magazine and Museum} (itself a merger of \textit{The Lady's Magazine} and a competitor) to become \textit{The Court Magazine and Monthly Critic}.}
\end{quote}

Nevertheless, the correct annotation is \texttt{NIL} since ``Court Magazine'' and ``La Belle Assemblée'' refer to distinct entities. This case illustrates how such ambiguities can confuse annotators.

Consider another example sentence from \benchmarkEL{}, extrapolated from an issue of the music periodical \textit{The Quarterly Musical Magazine And Review}, dating back to 1828:
\begin{example}\label{ex:court-of-vienna}
    \textit{Sommer, organist to the Court of Vienna; this is the greatest compliment which can be paid to his talents.}
\end{example}

In Example \ref{ex:court-of-vienna}, one annotator labeled ``Court of Vienna'' as \texttt{NIL}, while the other linked it to the ``Vienna State Opera'' \footnote{Available at the URLs \url{https://en.wikipedia.org/wiki/Vienna_State_Opera} and \url{https://www.Wikidata.org/wiki/Q209937}, last time accessed on August 12th, 2024.}.
The source of this ambiguity can be traced back again to information present in Wikipedia, which misled the annotator:

\begin{quote}
\textit{The Vienna State Opera [...] is an opera house and opera company based in Vienna, Austria [...]. The opera house was inaugurated as the ``Vienna Court Opera'' (Wiener Hofoper) in the presence of Emperor Franz Joseph I and Empress Elisabeth of Austria. It became known by its current name after establishing the First Austrian Republic in 1921.}
\end{quote}

After its construction, the opera house was known as the ``Vienna Court Opera''.
However, the construction happened after the date on which the periodical from which Example \ref{ex:court-of-vienna} is extracted was published, which is 1828, as it can be read in \benchmarkEL's metadata. Therefore, the correct annotation is \texttt{NIL}.

After calculating the IAA and identifying the discrepancy cases, we reconcile the output by manually aligning each problematic instance with the Annotation Guidelines. This ensures that the final version of the benchmark reports the correct cases.

\subsection{Statistics}
\label{subsub:stats_filtered}

\begin{table}[ht]
    \parbox{.65\linewidth}{
        \parbox{\linewidth}{
            \centering
            \caption{Documents' statistics for \benchmarkEL{} and HIPE-2020 test set.}
            \begin{tabular}{ccccc}
                \toprule
                Dataset & \#docs & \#sents & \#tokens & \makecell{\#tokens \\ per sentence \\ (average)} \\
                \midrule
                \benchmarkEL{} & $76$ & $875$ & $27,549$ & $31.5$\\
                HIPE-2020 & $46$ & $553$ & $16,635$ & $30$\\
                \bottomrule
            \end{tabular}
            \label{tab:MHERCLv01_HIPE2020_DatasetStats}
        }%
        \vspace{1em}
        \parbox{\linewidth}{
            \centering
            \caption{Named entities statistics for \benchmarkEL{} and HIPE-2020 test set.}
            \begin{tabular}{cccccc}
                \toprule
                Dataset & Count & \#mention & \#types & noisy & \texttt{NIL} \\
                \midrule
                \multirow{2}{*}{\benchmarkEL{}} & all & $2,370$ & $N/A$ & $0.14$ & $0.30$ \\
                                       & unique & $1,805$ & $58$ & $N/A$ & $0.38$ \\
                \midrule
                \multirow{2}{*}{HIPE-2020} & all & $449$ & $N/A$ & $0.05$ & $0.40$ \\
                                       & unique & $232$ & $5$ & $N/A$ & $0.54$ \\
                \bottomrule
            \end{tabular}
            \label{tab:MHERCLv01_HIPE2020_NamedEntitiesStats}
        }
    }%
    \hfill
    \parbox{.3\linewidth}{
        \caption{Top 10 named entity types occurring in \benchmarkEL{}}
        \begin{tabular}{ll}
        \toprule
        Type & \# \\
        \midrule
        \textsc{person}	& $1,253$\\
        \textsc{city}	& $262$\\
        \textsc{music}	& $187$\\
        \textsc{organization}	& $93$\\
        \textsc{work-of-art} & $85$\\
        \textsc{country}	& $80$\\
        \textsc{building}	& $52$\\
        \textsc{opera}	& $52$\\
        \textsc{theatre}	& $42$\\
        \textsc{worship-place}	& $41$\\
        \bottomrule
        \end{tabular}
        \label{tab:ne_types_stats_filtered}
    }
\end{table}

Table \ref{tab:MHERCLv01_HIPE2020_DatasetStats} and Table \ref{tab:MHERCLv01_HIPE2020_NamedEntitiesStats} report the main statistics for \benchmarkEL{}, and compare it to the most similar resource available for HEL, HIPE-2020.
\benchmarkEL{} is composed of $875$ sentences, extrapolated from $76$ historical periodicals.
They include $1,805$ unique named entities belonging to $58$ different types. As it is possible to observe from the comparison, \benchmarkEL{} is larger than HIPE-2020 English test set with respect to the number of sentences included and the annotated entities. Also, it is more varied regarding source documents and the named entity types considered.

As reported in Table \ref{tab:MHERCLv01_HIPE2020_NamedEntitiesStats}, $30\%$ of the total of annotated named entity mentions could not be linked to any Wikidata entity, which resulted in \texttt{NIL} annotations. This particular aspect makes \benchmarkEL{} a challenging dataset from a linking standpoint, as the model must be able to predict an entity conservatively - i.e., predicting \texttt{NIL} - or it is impossible to obtain an accuracy higher than $70\%$.
Moreover, in Table \ref{tab:MHERCLv01_HIPE2020_NamedEntitiesStats}, the amount of noisy entities is reported.
These are the annotations impacted by errors due to OCR procedure on the original periodicals.

Table \ref{tab:ne_types_stats_filtered} reports the top 10 named entity types occurring in \benchmarkEL{} in order of frequency. The types distribution reveals the benchmark's specificity to the music domain due to the presence of domain-specific types such as \textsc{music}, \textsc{opera}, and \textsc{work-of-art}. A full description of the types used for NEs classification is provided in Appendix \ref{app:A}.

\subsection{Popularity study} \label{sec:popularity}

\begin{figure}[ht]
    \centering
    \begin{subfigure}[t]{0.4\textwidth}
        \centering
        \includegraphics[width=\textwidth]{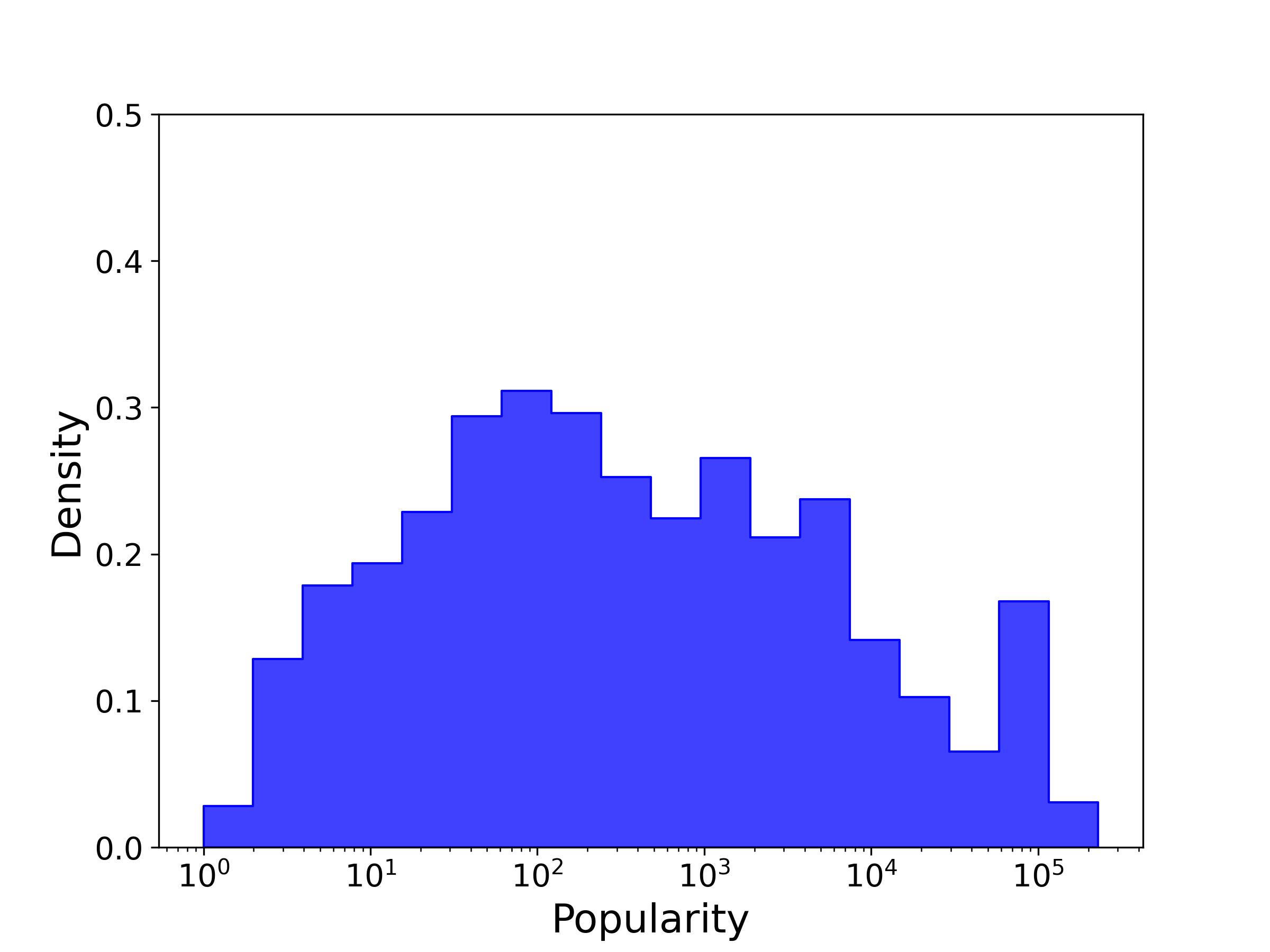}
        \caption{\benchmarkEL{}}
        \label{fig:mhercl_popdist}
    \end{subfigure}
    \begin{subfigure}[t]{0.4\textwidth}
        \centering
        \includegraphics[width=\textwidth]{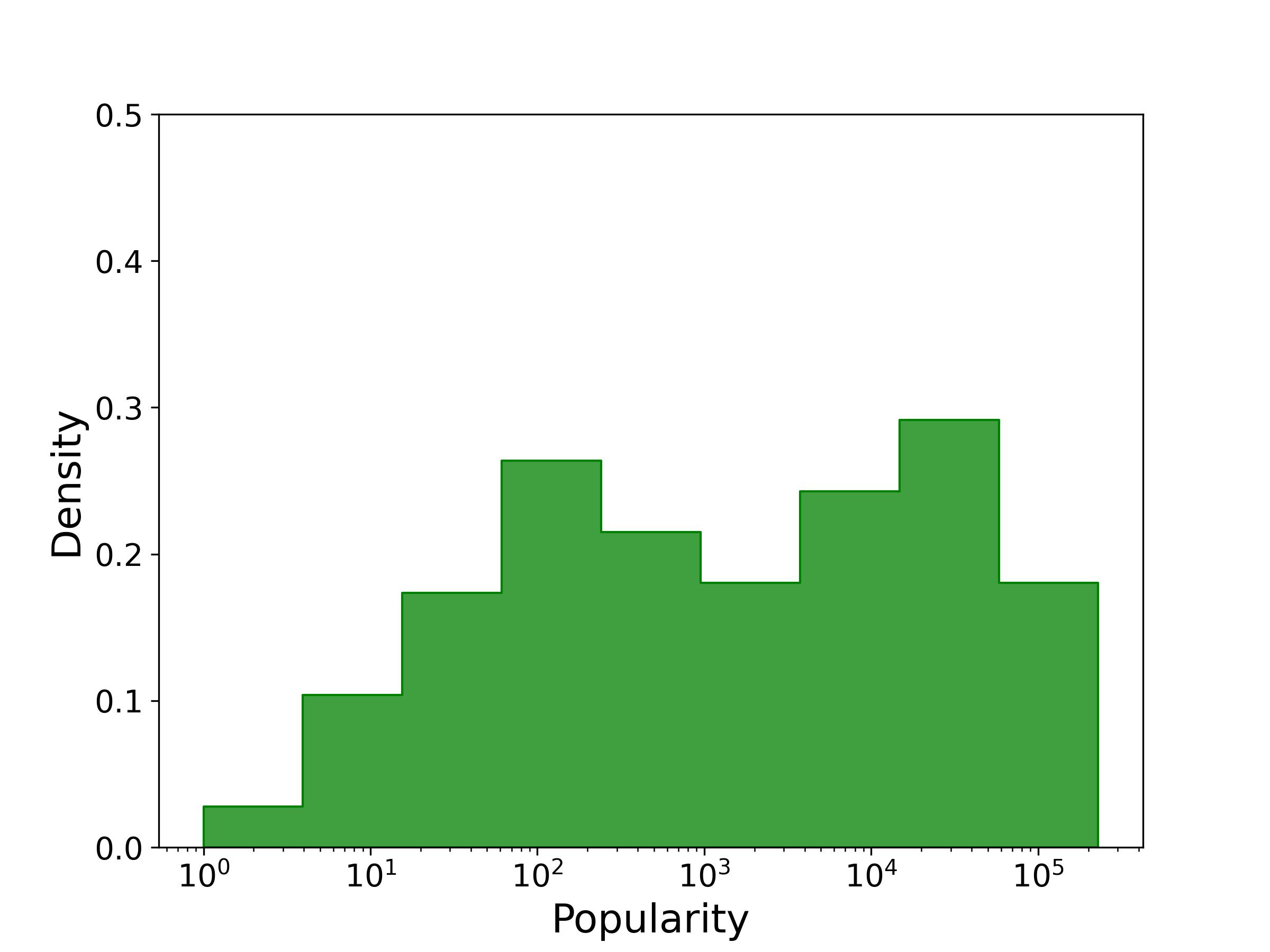}
        \caption{HIPE-2020}
        \label{fig:clef_popdist}
    \end{subfigure}
    
    \begin{subfigure}[t]{0.4\textwidth}
        \centering
        \includegraphics[width=\textwidth]{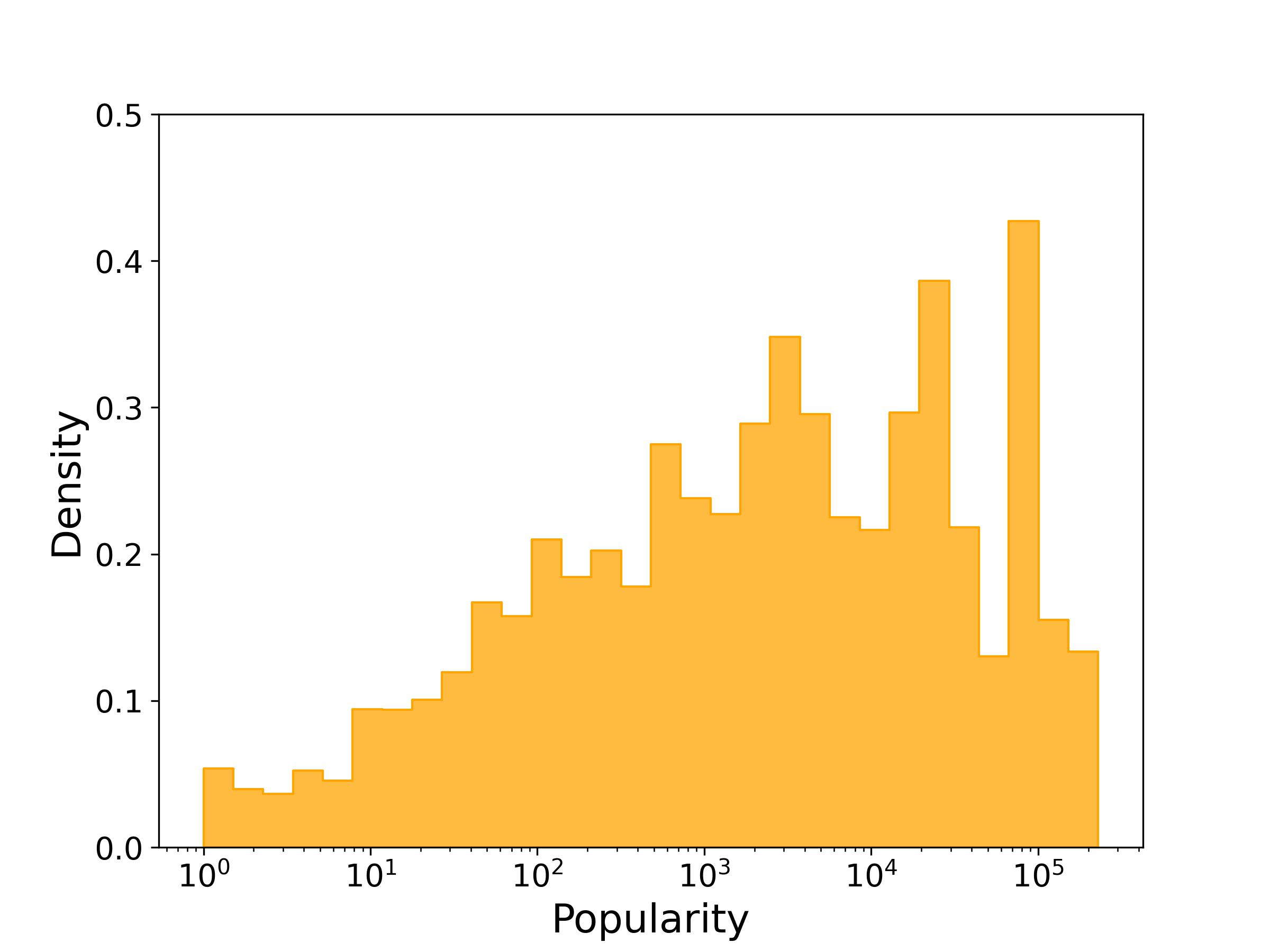}
        \caption{AIDA CoNLL-YAGO}
        \label{fig:aida_popdist}
    \end{subfigure}
    \begin{subfigure}[t]{0.4\textwidth}
        \centering
        \includegraphics[width=\textwidth]{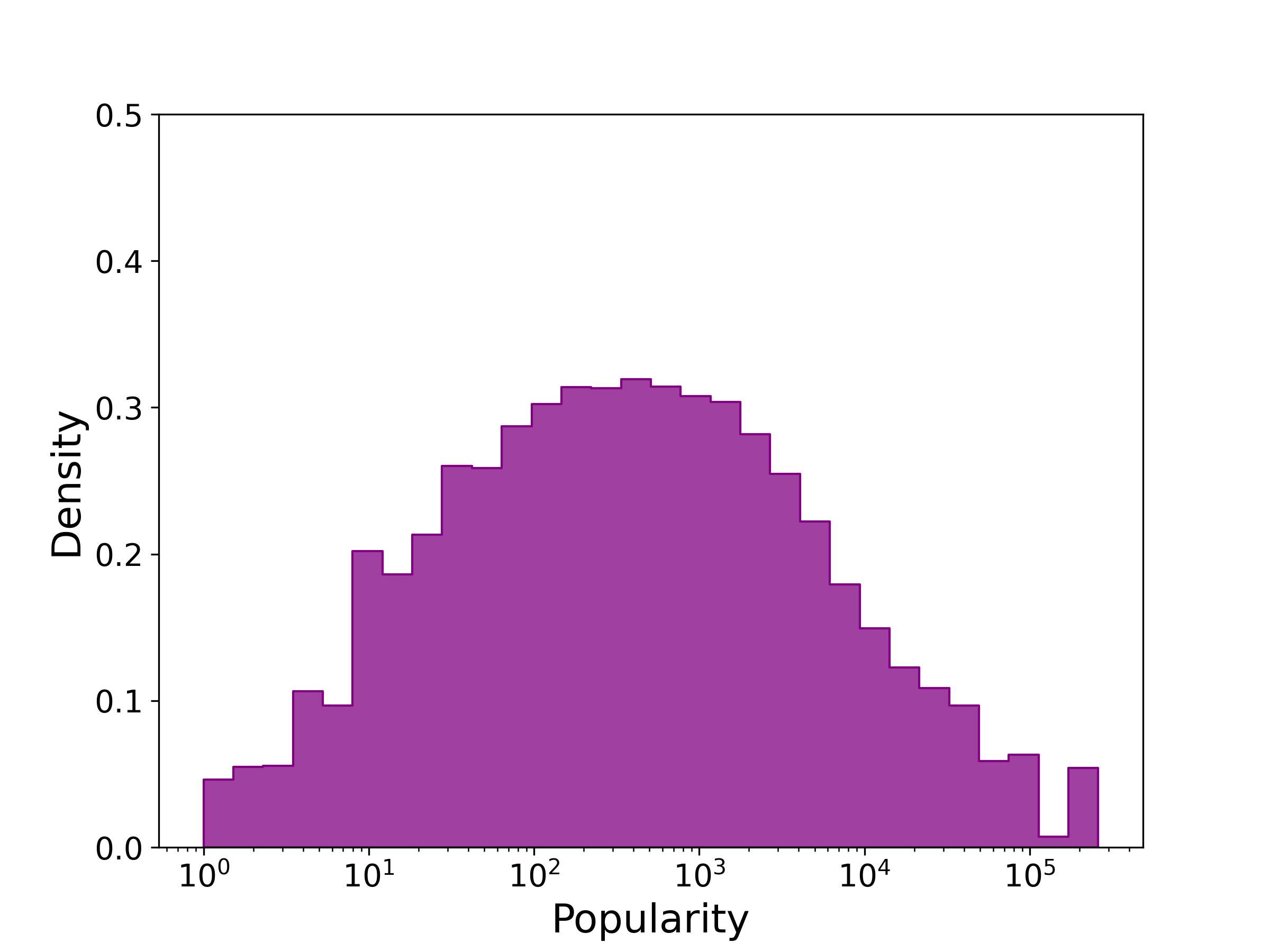}
        \caption{BLINK's training set}
        \label{fig:blinkts_popdist}
    \end{subfigure}
    
    \caption{Figures showing the distribution of named entity popularity in \benchmarkEL{} (\ref{fig:mhercl_popdist}), HIPE-2020 (\ref{fig:clef_popdist}), AIDA-COnLL-YAGO (\ref{fig:aida_popdist}) benchmarks and BLINK's training dataset (\ref{fig:blinkts_popdist}). Each figure contains a histogram showing the density of entities' popularity levels: the higher the density, the more common the entities with the corresponding popularity value. Popularity is computed as the frequency of occurrence of each named entity's QID as an internal link in Wikipedia. \texttt{NIL} are excluded in \benchmarkEL{} and HIPE-2020.}
    \label{fig:ent_pop_dist_across4datasets}
\end{figure}

\begin{figure}[ht]
    \centering
    \includegraphics[width=0.60\textwidth]{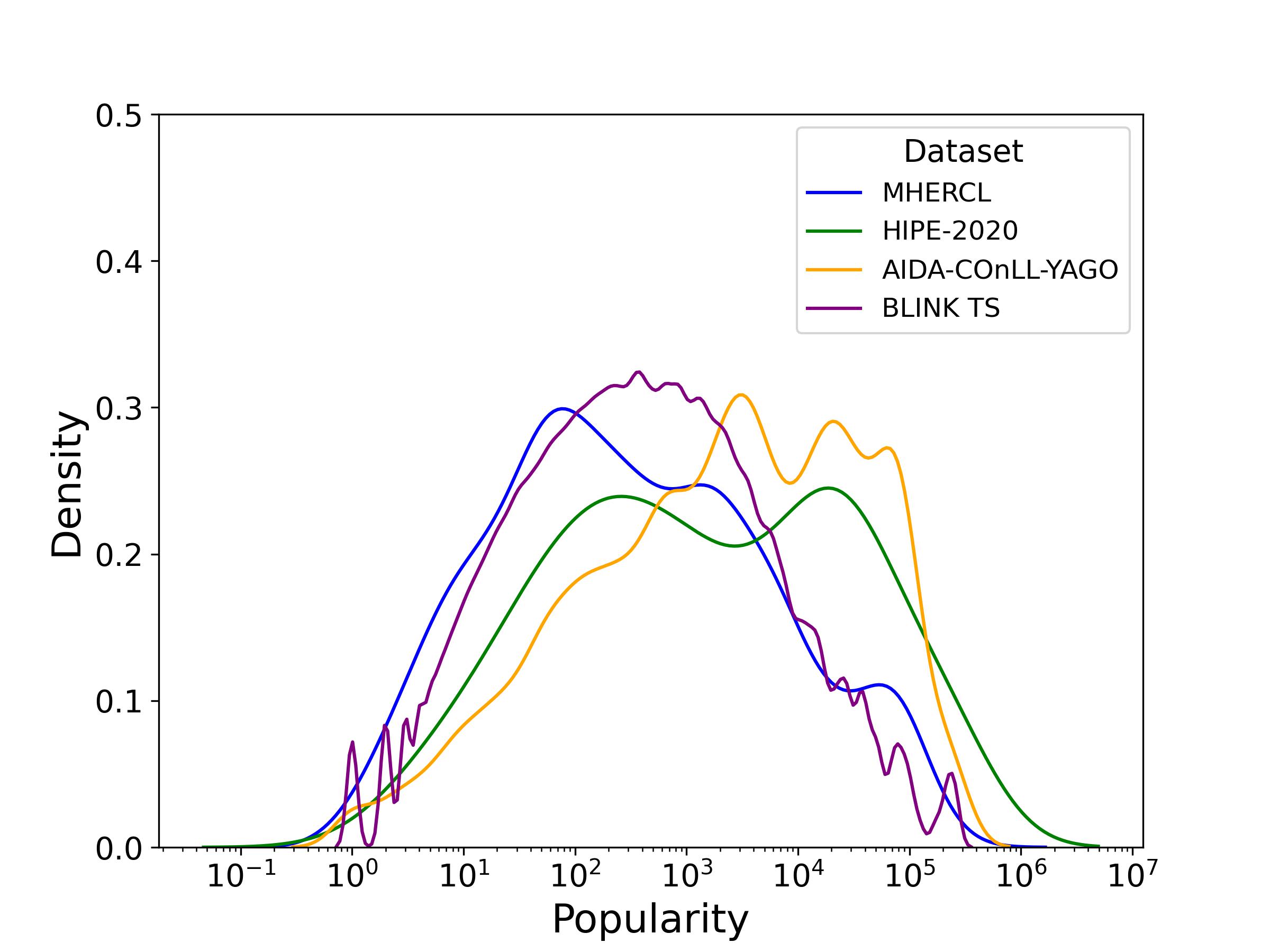}
    \caption{Kernel Density Estimation (KDE) plot comparing the smoothed density functions of named entities popularity across the \benchmarkEL{}, HIPE-2020, AIDA CoNLL-YAGO and BLINK's Training Set (TS) datasets. For \benchmarkEL{} and HIPE-2020, only entities with a valid QID are considered, \texttt{NIL} entities being excluded from the analysis.}
    \label{fig:kde_mhercl_clef_aida}
\end{figure}

Numerous studies tackle popularity bias in knowledge-intensive downstream tasks \citep{kandpalLargeLanguageModels2023, mallen-etal-2023-trust, sunHeadtoTailHowKnowledgeable2024}, including EL \citep{ilievski-etal-2018-systematic, ayoola-etal-2022-improving}. To show the extent to which \benchmarkEL{} may contribute to the advancements of research for EL on long-tail (i.e., low popularity) entities, we analyze the popularity distribution of its named entities, and we compare it to: \begin{enumerate*}[label=(\arabic*)] \item the HIPE-2020 English test set \citep{HIPE2020Overview,HIPE2020ExtOverview}; \item the AIDA CoNLL-YAGO EL benchmark (English) \citep{hoffart-etal-2011-robust}, \item BLINK's Training Set (TS), which consists of a 9M examples subset of the May 2019 English Wikipedia dump.\end{enumerate*}
AIDA CoNLL-YAGO and BLINK's TS are taken in their KILT \citep{petroni-etal-2021-kilt} release.
As a proxy for popularity, we calculate each named entity's QID frequency of occurrence as an internal link in Wikipedia\footnote{For popularity calculation, we used the Wikipedia dump \texttt{enwiki-20220120} dump.}. \texttt{NIL} entities in \benchmarkEL{} and HIPE-2020 are excluded from the analysis.

Aligning with \citet{mallen-etal-2023-trust}, we illustrate in Figure \ref{fig:ent_pop_dist_across4datasets} the distribution of named entities' popularity across the three datasets chosen.  As can be observed in Figure \ref{fig:mhercl_popdist}, \benchmarkEL{} displays higher density levels at lower popularity values. This means that, on average, an entity that is randomly drawn from \benchmarkEL{} is more likely to be associated with low popularity values (between $10$ and $10^3$).
On the contrary, as shown by Figure \ref{fig:aida_popdist}, AIDA CoNLL-YAGO named entities exhibit a higher density at higher popularity values, showing that traditional benchmarks mostly focus on highly represented entities. Despite its historical focus, as shown in Figure \ref{fig:clef_popdist}, HIPE-2020 presents a distribution that resembles the sum of two normal distributions centred at moderately popular entities ($\approx 10^2$) and highly popular entities ($\geq 10^4$). It is hence biased towards popular entities, similarly to AIDA CoNLL-YAGO, but also maintains a considerable representation of moderately popular entities, which are the prevalent type of entities in training datasets (see BLINK's TS distribution of Figure \ref{fig:blinkts_popdist}). 
Indeed, from Figure \ref{fig:blinkts_popdist}, it can be seen how BLINK's TS, which is derived from Wikipedia, has a wide coverage of highly popular entities as well as low-popular ones. This proves that models processing historical documents can benefit from training on general-purpose datasets, as their popularity distribution includes low-popularity entities as well.

As can be observed in Figure \ref{fig:kde_mhercl_clef_aida} reporting the Kernel Density Estimation (KDE) plots, all three datasets share a common distribution of entities within the mid-range popularity values ($10^1$ to $10^4$). Despite this overlap, the plot underlines that HIPE-2020 and \benchmarkEL{} are complementary datasets for historical documents since they focus on a similar domain but display different data distributions. From a popularity point of view, HIPE-2020 is more similar to AIDA CoNLL-YAGO, while \benchmarkEL{} characterizes itself as a benchmark complementary to standard ones -- e.g. AIDA CoNLL-YAGO. This further highlights the novelty and importance of our dataset for studies focusing on low-popularity entities, extending its value beyond HEL to broader applications in EL research.

\section{Models}
\label{sec:models}

Due to the limitations of current entity linkers in HEL discussed in Section \ref{sec:hel-introduction}, models tailored to the unique requirements of historical texts are needed. In this section, we present two methods based on two different architectures to enhance entity linkers for HEL.

The first method, Entity Linking Dynamics (\modelEL{}, Section \ref{sub:eld}), is an unsupervised model based on game-theoretic principles. It bypasses explicit training on labelled sentences, making it suitable for low-popularity entities typical of historical documents, and explicitly handles \texttt{NIL} predictions. The second method, Constrained-BLINK (\modelCBLINK{}, Section \ref{sub:c-blink}) extends BLINK by adding type and temporal plausibility constraints during the candidate retrieval phase and by including \texttt{NIL} prediction without the need to re-train the model.

\subsection{Entity Linking Dynamics (ELD)}\label{sub:eld}
\subsubsection{Theoretical Foundation}
\modelEL{}'s approach to EL assumes that the disambiguation process underlying this task has to be coherent with the context in which entities appear, respecting time and logical constraints.
It is grounded on relaxation labelling processes (RLP) \citep{4767390}. 
This class of algorithms produce consistent labelling of the data thanks to the use of constraints among labels (entities in the context of EL) designed to make the label assigned to an object consistent with the labels assigned to the other objects in the same batch (sentence or paragraph in the context of EL). 
RLPs have been used successfully in different fields, including computer vision \citep{DBLP:journals/jmiv/Pelillo97}, Natural Language Processing (NLP) \citep{tripodi-navigli-2019-game}, and graph processing \citep{ROTABULO2011984}.
One possible formulation of RLPs is in terms of game theory \citep{DBLP:journals/orl/MillerZ91}.
In this formulation, objects to be labelled are interpreted as the players of a non-cooperative game \citep{nash1951non}, and the constraints on labels are the strategies that the players can select to play the game. 

\begin{table}
    \captionsetup{width=\textwidth}
    \caption{Payoff matrix for the game of \emph{matching pennies}. The number on the left of each cell of the matrix indicates the payoff obtained by player 1, and the number on the right indicates the payoff of player 2.
    Player 1 plays according to the rows of the matrix, and player 2 with the columns.}
    \centering
    \begin{tabular}{ll | lr | lr }
        \toprule
        & & \multicolumn{4}{c}{PLAYER 2}\\
        & & \multicolumn{2}{c}{heads} & \multicolumn{2}{c}{tails} \\ \cmidrule(lr){3-6}
        \multirow{2}{*}{PLAYER 1} &
        heads & 1 & -1 & -1 & 1 \\ \cmidrule(lr){2-6}
        & tails &  -1 & 1 & 1 & -1 \\
        \bottomrule
    \end{tabular}
    
    \label{tab:payoff_mat}
\end{table}

In game theory, particularly in non-cooperative games, the players are considered rational and select the strategy that maximizes a predefined payoff.
The selection is performed considering the strategies a co-player can select in response to a strategy.
The payoff of different strategy combinations is stored in a payoff matrix that defines the game.
For example, consider a two-player game, i.e., \emph{matching pennies}, where a coin is flipped, and the strategies of the players correspond to the selection of one face of the coin (head or tail).
Player 1 wins if both players select the same face of the coin, while player 2 wins if they choose different faces.
If both players select heads, player 1 will have a positive payoff, while player 2 will have a negative payoff.
The payoff matrix, reported in Table \ref{tab:payoff_mat}, defines the payoff of both players with respect to their gaming strategy - i.e. if they chose heads or tails.
As a consequence, the outcome of a game does not depend on the strategy selected by a single player but on the combination of the strategies selected by the players involved in the game.
An equilibrium state for the game described above, i.e. the optimal strategy that both players should follow, is to use a mixed strategy in which heads and tails are played with the same probability.
In this way, it is difficult for a player to guess the strategy that the other player will use.
In this case, we say that the players are using \emph{mixed strategies}, that are probability distributions over pure strategies.

A mixed strategy for a player $i$ is defined as a probability vector $\mathbf{x_i}=(x_{1},\ldots,x_{m_i} )$, such that $x_{j} \geq 0$ and $\sum x_j = 1$.
Each mixed strategy corresponds to a point in the simplex $\Delta_m$, whose corners correspond to pure strategies.
The intuition is that players randomize over strategies according to the probabilities in $\mathbf{x_i}$.
Each mixed strategy profile lives in the mixed strategy space of the game, given by the Cartesian product $\Theta = \Delta_{m_1} \times \Delta_{m_2} \times \dots \times \Delta_{m_n}$.
In a \emph{two-player game}, a strategy profile can be defined as a pair $(\mathbf{x}_i,\mathbf{x}_j)$.
The expected payoff for this strategy profile is computed as:
\begin{equation*}
u(\mathbf{x}_i,\mathbf{x}_j)=\mathbf{x}_i^T \cdot A_{ij} \mathbf{x}_j
\end{equation*}
\noindent where $A_{ij}$ is the $m_i \times m_j$ payoff matrix between player $i$ and $j$.
In evolutionary game theory \citep{weibull1997evolutionary}, the games are played repeatedly, and the players update their mixed strategy distributions over time until no player can improve the payoff obtained with the current mixed strategy.
This situation corresponds to the equilibrium of the system.
This dynamic process allows to perform consistent labeling of the data \citep{DBLP:journals/orl/MillerZ91}, and from this aspect, we assign a name to our system, i.e. Entity Linking Dynamics.

In \modelEL{}, the EL task is framed as a game in which each player is a mention that has to select an entity to be linked to.
The strategy of the game corresponds to the entities to which a mention can be linked towards a KB.
The Entity Linking games are formulated by modeling the ability of each player to select the most consistent strategy, i.e. the strategy that grants the highest payoff. 
This is done using a KB to select all the possible entities to which each mention can be associated.
These entities are interpreted as the strategies that the player can select.
The linking process is performed jointly with Word Sense Disambiguation (WSD), i.e., associating each content word in a sentence with its meaning as encoded in a KB such as WordNet \citep{Miller1995}.
This allows the exploitation of contextual information to perform consistent data labelling, disambiguate entities, and consider nouns and verbs around them.
ELD leverages historical context as implicit knowledge, which helps the model to make connections between entities and assign a higher payoff to entities that are related. This allows to perform consistent labeling of the data, which means disambiguating an entity taking into account the relations that it has with all other entities in the text. Related entities could be entities that existed in the same period or entities that can be associated with the same domain, such as artists and musicians in the context of MERCHL.
This approach enables the model to capture nuanced relationships that may not be explicitly defined in labelled training datasets, particularly when disambiguating entities individually rather than simultaneously, as is done in ELD.

\subsubsection{ELD model}
In \modelEL{}, an input sentence is pre-processed with a tokeniser, a part-of-speech tagger and a named entity recognition tagger to obtain a set of annotations $T=((t_1, pos_1, ner_1),(t_2, pos_2, ner_2),\dots, (t_n, pos_n, ner_n))$.
Additionally, the text is fed into a Pre-trained Language Model to obtain a contextualised representation of each token $E=(\mathbf{e}_1, \mathbf{e}_2, \dots, \mathbf{e}_n)$.
The subtoken representations of $E$ are averaged to match the tokenisation in $T$.
The result is a feature vector for each token in $T$, which results in a $n \times w$ matrix $W$ in which only the $n$ nouns, verbs, adjectives, adverbs and named entities in $T$ are maintained and $w$ refers to the dimension of the feature vectors.
These $n$ tokens are the players of the game implemented in \modelEL{}. Based on the pairwise similarities computed using $W$, \modelEL{} weights each player's influence on each other such that similar players have a reciprocal influence on their strategy selection.
For example, in the sentence:

\begin{quote}
    \textit{Abe Lincoln played the guitar in the Dinosaurs band for 20 years.}
\end{quote}

\noindent the word \textit{guitar} influences the mention \textit{Abe Lincoln} in the entity \textit{Abe\_Lincoln\_(guitarist)} rather than \textit{Abe\_Lincoln\_(politician)}.
This information is obtained by constructing a payoff matrix that encodes the similarity between the word \emph{guitar} and the entities \textit{Abe\_Lincoln\_(guitarist)} and \textit{Abe\_Lincoln\_(politician)}. 

The candidate generation phase is performed by selecting the possible senses/entities to which a word/mention can refer.
We rely on two different KBs: WordNet for the selection of the senses of nouns, verbs, adjectives and adverbs, and Wikidata for the selection of entities connected to a mention marked as named entity.

In \modelEL{}, the strategy space is a probability distribution encoded in an $n \times m$ matrix $X$ where $n$ is the number of players (the mentions in the sentence) and $m$ the entities to link to.
It can be defined using a prior, if available, or through a uniform distribution. 
The payoff matrix is defined by computing embeddings of each entity and sense such that they all lie in the same latent space - i.e. they are compatible representations.
In particular, sense embeddings are obtained using Ares embeddings \citep{scarlini-etal-2020-contexts}, employing a multilingual language model to embed text annotated with WSD. In contrast, entity embeddings are obtained using the same multilingual language model used by Ares, i.e., mBERT \citep{devlin-etal-2019-bert}. 
The $m \times m$ matrix $Z$, represents the payoff matrix of the game.
It is obtained by computing the pairwise similarity between entity embeddings.
This formulation allows to guide the disambiguation dynamics towards the senses of an entity that are more similar to it.
In the example above, the embedding \textit{guitar} has higher contextual similarity to the entity \textit{Abe\_Lincoln\_(guitarist)} than to other possible entities.

\modelEL{} exploits the matrix $Z$ by relying on the replicator dynamic equation \citep{TAYLOR1978145} where the payoff of strategy $h$ for player $i$ is calculated as:
\begin{equation}
    \label{eq:singlePayoff_}%
    u(x_i^h) = x_i^h \cdot \sum_{j=1}^{n_i}(A_{ij}Z \mathbf{x}_j)^h 
\end{equation}%
\noindent where $x_i^h$ is the probability of strategy $h$ for player $i$, according to the matrix $X$, $\mathbf{x}_j$ is the embedding of the entity $j$ and $n_i$ are the neighbours of player $i$, defined by the adjacency matrix $A$. Informally, $A$ encodes which senses are neighbours of the mention $i$ in the input sentence.
The entity with a higher payoff with the target mention is selected for the final link.

\subsection{C-BLINK}\label{sub:c-blink}

Constrained-BLINK (\modelCBLINK{}) extends BLINK by filtering entities based on type and temporal plausibility during candidate retrieval. Type and date information are sourced from an external KB (Wikidata).

In HEL, a \textit{plausible} candidate is one whose type loosely aligns with the manually annotated named entity type and whose time information precedes the document's date. For instance, consider the following sentence from \benchmarkEL{}, originally found in a 1824 issue of \textit{The Quarterly Musical Magazine And Review}:

\begin{quote}
    \textit{Fabio Constantini flourished about the year 1630, and ultimately became maestro at the chapel of Loretto.}
\end{quote}
    
\noindent BLINK links \textit{Fabio Constantini} to \textit{Claudio Constantini}\footnote{Available at \url{https://en.wikipedia.org/wiki/Claudio_Constantini} and \url{https://www.wikidata.org/wiki/Q5129347}, last accessed on August 12th, 2024.}, which is implausible, as the document dates to 1824 and \textit{Claudio Constantini} was born in 1983.

Following the approach of \citet{tedeschi-etal-2021-named-entity}, we limit the candidates considered by BLINK's to \textit{plausible} entities. Specifically, \modelCBLINK{} extends BLINK's candidate retrieval by filtering its \textit{bi-encoder} module using a collection of boolean functions $\Phi$. Each boolean function $\phi \in \Phi$ checks whether an entity $e$ is plausible for a mention $m$. For example, it is possible to implement the approach of \citep{tedeschi-etal-2021-named-entity} by defining $\phi_t \in \Phi$ as that function that evaluates to true if the NEC class predicted for the mention $m$ matches the class predicted for the entity $e$.

Moreover, we define $\phi_d \in \Phi$ as the function that evaluates to true only if the date associated with the entity $e$ precedes the date associated with the mention $m$. In the example above, $\phi(\textit{Fabio Constantini}, \textit{Claudio Constantini})$ is false since \textit{Fabio Constantini} is mentioned on a document issued in 1824 and \textit{Claudio Constantini} was born in 1983.

In practice, the integration of such constraints in \modelCBLINK{} is implemented by directly annihilating the similarity scores computed by the bi-encoder for implausible entities. Our approach is agnostic to the choice of the plausibility function: $\phi \in \Phi$ can be heuristic, like $\phi_d$ or any arbitrary function, including a neural network. In the latter case, $m$ and $s$ are represented by their dense encodings from the bi-encoder. This allows for the straightforward reuse of pre-trained models without any additional training requirements. 

The plausibility constraints avoid the prediction of logically inconsistent entities, but it does not directly address \texttt{NIL} predictions for entities that are not present in our reference KB. Since all the candidates considered by the model are plausible, we assume that if the correct entity is present in the KB, then it must be the one with the highest similarity with respect to the document's mention. Note that this assumption directly aligns with retrieval-based models since they are trained to maximise the similarity between the document's dense representation and its associated entity. Our \texttt{NIL} prediction approach leverages this assumption. If there is no document in the pool of candidates whose similarity is significantly higher than the rest with respect to the input mention (according to the heuristics described later), we infer the absence of a confident link and predict \texttt{NIL}.

As discussed in Section \ref{sec:related}, different approaches have been proposed to perform \texttt{NIL} prediction. We design an extensive set of heuristics and classifiers to assess their effectiveness on this task, acting on three main aspects: \begin{enumerate*}[label=(\arabic*)]
\item the candidates' scores, 
\item the string similarity between the superficial mention found in the benchmarks and the predicted entity, 
\item Machine learning-based methods.
\end{enumerate*}

\begin{table}[ht]
    \centering
    \caption{Description of the statistical heuristics. $S = [s_0, \cdots, s_n]$ is used for the scores provided by the model to the entity that can be linked, $\tau \in [0, 1]$.}
    \begin{tabular}{ccm{6cm}}
        \toprule
        Name & Condition & Description \\ \midrule
         Fixed threshold & $s_0 < \tau$ & The top score should be higher than $\tau$ \\
         Deviation from top & $\frac{s_0 - s_1}{(s_0 + s_1) / 2} < \tau$ & The percentage difference between the two top scores should be higher than $\tau$ \\
         Deviation from median & $\frac{s_0 - \overline{S}}{(s_0 + \overline{S}) / 2} < \tau$ & The percentage difference between the top score and the median of the scores ($\overline{S}$) should be higher than $\tau$ \\
         Deviation from mean & $\frac{s_0 - \mu(S)}{(s_0 + \mu(S)) / 2} < \tau$ & The percentage difference between the top score and the mean of the scores ($\overline{S}$) should be higher than $\tau$ \\ \bottomrule
    \end{tabular}
    \label{tab:nil_heuristics}
\end{table}

Table \ref{tab:nil_heuristics} describes the heuristics that act on the scores provided by the model. For the string similarity heuristics, we experiment with Levenshtein distance, Jaccard distance, and Hamming distance \citep{wang2020similaritymeasures}. All those methods involve the use of a threshold value. For machine learning models, we experimented with Support Vector Machines (SVM), logistic regression, and decision trees.

We extensively test \modelCBLINK{} and report a fine-grained description of the effects of $\phi_d$, $\phi_t$ and \texttt{NIL} heuristics on its performance in Section \ref{sec:experiments}.

\section{Experiments}
\label{sec:experiments}

\begin{table}[!htbp]
    \captionsetup{justification=justified,singlelinecheck=false}
    \caption{Baselines, \modelEL{} and \modelCBLINK{} results on HIPE2020 and \benchmarkEL{}. The highest F1 score is in bold. The application of which constraint (see Sections \ref{sec:time-c} and \ref{sec:type-c}), and NIL heuristics (see Table \ref{tab:nil_heuristics}) is reported alongside their threshold \textit{$\tau$}, if needed. \textit{Always \texttt{NIL}} baseline is a dummy model that always predicts \texttt{NIL} for each mention. BLINK$^\dagger$ and \modelCBLINK{}$^\dagger$ do not rely on the re-ranking phase. \modelEL{}$^\dagger$ does not use the replicator dynamics but selects the entity that is most similar to the mention. Models with * do not receive the gold named entity mention in the input.}
    \begin{tabular}{lccccc}
    \toprule
    Model & $\Phi$ & \texttt{NIL} heuristic & $\tau$ & HIPE-2020 & \benchmarkEL{} \\ \midrule
    Always \texttt{NIL} & $-$ & $-$ & $-$ & $0.43$ & $0.31$ \\

    \citep[REL]{REL2020} & $-$ &  $-$  & $-$ & $0.16$ & $0.10$ \\
    
    \citep[GENRE]{decao2021autoregressive} & $-$ & $-$ & $-$ & $0.30$ & $0.41$ \\
    
    \citep[mGENRE]{de-cao-etal-2022-multilingual} & $-$ & $-$ & $-$ & $0.31$ & $0.47$ \\

    \citep[CLOCQ*]{Christmann2022clocq} & $-$ & $-$ & $-$ & $0.10$ & $0.16$ \\
    
    \citep[ReFinED*]{ayoola-etal-2022-improving} & $-$ & \texttt{NIL} in candidates & $-$ & $0.46$ & $0.52$ \\

    \citep[ExtEnD*]{barba-etal-2022-extend} & $-$ & $-$ & $-$ & $0.04$ & $0.25$ \\
    
    \citep[ReLiK*]{Orlando2024} & $-$ & $-$ & $-$ & $0.07$ & $0.31$ \\

    \citep[BLINK]{wu-etal-2020-scalable} & $-$ & $-$ & $-$ & $0.33$ & $0.53$ \\
    \citep[BLINK$^\dagger$]{wu-etal-2020-scalable} & $-$ & $-$ & $-$ & $0.34$ & $0.52$ \\ 

    \textcolor{orange}{LLAMA 3.3 70B} & $-$ & \texttt{NIL} allowed by prompt & $-$ & $0.60$ & $0.61$ \\ 
    \textcolor{orange}{GPT-4 o1-mini} & $-$ & \texttt{NIL} allowed by prompt & $-$ & $0.68$ & $0.60$ \\ 
    
    \midrule


    \modelEL{} & $-$ & \texttt{NIL} in candidates & $-$ & $0.62$ & $0.56$\\
    \modelEL{} & $\phi_d$ & \texttt{NIL} in candidates & $-$ & $0.63$ &  $0.57$\\
    \modelEL{} & $\phi_t$ & \texttt{NIL} in candidates & $-$ & $0.62$ & $0.57$\\
    \modelEL{} & $\{ \phi_d, \phi_t \}$ & \texttt{NIL} in candidates & $-$ & $0.62$ & \textbf{$0.58$}\\

    \modelEL{}$^\dagger$ & $-$ & \texttt{NIL} in candidates & $-$ & $0.46$ & $0.49$\\
    \modelEL{}$^\dagger$ & $\phi_d$ & \texttt{NIL} in candidates & $-$ & $0.47$ & $0.45$\\
    \modelEL{}$^\dagger$ & $\phi_t$ & \texttt{NIL} in candidates & $-$ & $0.48$ & $0.49$\\
    \modelEL{}$^\dagger$ & $\{ \phi_d, \phi_t \}$ & \texttt{NIL} in candidates & $-$ & $0.49$ & $0.49$\\


    \modelCBLINK{} & $\phi_d$ & $-$ & $-$ & $0.34$ & $0.54$ \\
    \phantom{\modelCBLINK{}} + & $\phi_t$ & $-$ & $-$ & $0.34$ & $0.54$ \\
    \phantom{\modelCBLINK{}} + & $\{ \phi_d, \phi_t \}$ & $-$ & $-$ & $0.35$ & $0.54$ \\
    \phantom{\modelCBLINK{}} + & $\{ \phi_d, \phi_t \}$ & Fixed threshold & $0.443$ & $0.66$ & $\textbf{0.72}$ \\
    \phantom{\modelCBLINK{}} + & $\{ \phi_d, \phi_t \}$ & Deviation from top & $0.003$ & $0.60$ & $0.63$ \\
    \phantom{\modelCBLINK{}} + & $\{ \phi_d, \phi_t \}$ &  Deviation from median & $0.003$ & $0.60$ & $0.63$ \\
    \phantom{\modelCBLINK{}} + & $\{ \phi_d, \phi_t \}$ &  Deviation from mean & $0.000$ & $0.57$ & $0.59$ \\
    \phantom{\modelCBLINK{}} + & $\{ \phi_d, \phi_t \}$ &  Levenshtein & $0.788$ & $0.60$ & $0.49$ \\
    \phantom{\modelCBLINK{}} + & $\{ \phi_d, \phi_t \}$ &  Jaccard & $0.613$ & $0.57$ & $0.53$ \\
    \phantom{\modelCBLINK{}} + & $\{ \phi_d, \phi_t \}$ &  Hamming & $0.402$ & $0.58$ & $0.48$ \\
    \phantom{\modelCBLINK{}} + & $\{ \phi_d, \phi_t \}$ &  SVM & $-$ & $0.65$ & $0.69$ \\
    \phantom{\modelCBLINK{}} + & $\{ \phi_d, \phi_t \}$ &  Logistic regression & $-$ & $0.64$ & $0.72$ \\
    \phantom{\modelCBLINK{}} + & $\{ \phi_d, \phi_t \}$ &  Decision Tree & $-$ & $\textbf{0.77}$ & $0.59$ \\ 

    \modelCBLINK{}$^\dagger$ & $\phi_d$ & $-$ & $-$ & $0.34$ & $0.53$ \\
    \phantom{\modelCBLINK{}$^\dagger$} + & $\phi_t$ & $-$ & $-$ & $0.34$ & $0.53$ \\
    \phantom{\modelCBLINK{}$^\dagger$} + & $\{ \phi_d, \phi_t \}$ & $-$ & $-$ & $0.34$ & $0.54$ \\
    \phantom{\modelCBLINK{}$^\dagger$} + & $\{ \phi_d, \phi_t \}$ & Fixed threshold & $0.626$ & $0.54$ & $0.49$ \\
    \phantom{\modelCBLINK{}$^\dagger$} + & $\{ \phi_d, \phi_t \}$ & Deviation from top & $0.011$ & $0.64$ & $0.72$ \\
    \phantom{\modelCBLINK{}$^\dagger$} + & $\{ \phi_d, \phi_t \}$ & Deviation from median & $0.025$ & $0.67$ & $0.74$ \\
    \phantom{\modelCBLINK{}$^\dagger$} + & $\{ \phi_d, \phi_t \}$ & Deviation from mean & $0.022$ & $0.67$ & $\textbf{0.76}$ \\
    \phantom{\modelCBLINK{}$^\dagger$} + & $\{ \phi_d, \phi_t \}$ & Levenshtein & $0.733$ & $0.56$ & $0.50$ \\
    \phantom{\modelCBLINK{}$^\dagger$} + & $\{ \phi_d, \phi_t \}$ & Jaccard & $0.613$ & $0.56$ & $0.53$ \\
    \phantom{\modelCBLINK{}$^\dagger$} + & $\{ \phi_d, \phi_t \}$ & Hamming & $0.467$ & $0.57$ & $0.47$ \\
    \phantom{\modelCBLINK{}$^\dagger$} + & $\{ \phi_d, \phi_t \}$ & SVM & $-$ & $0.66$ & $0.71$ \\
    \phantom{\modelCBLINK{}$^\dagger$} + & $\{ \phi_d, \phi_t \}$ & Logistic Regression & $-$ & $0.65$ & $0.74$ \\
    \phantom{\modelCBLINK{}$^\dagger$} + & $\{ \phi_d, \phi_t \}$ & Decision Tree & $-$ & $\textbf{0.76}$ & $0.63$ \\ 
    
    \bottomrule
    \end{tabular}
    \label{tab:MHERCL_results}
\end{table}

In this section, we detail our experimental setup designed to evaluate the performance of our HEL models. We conduct assessments of \modelEL{} and \modelCBLINK{} across both \benchmarkEL{} and HIPE-2020, comparing them against established baselines. These baselines encompass publicly accessible retrieval-based and generative models: REL \citep{REL2020}, BLINK \citep{wu-etal-2020-scalable}, GENRE \citep{decao2021autoregressive}, mGENRE \citep{de-cao-etal-2022-multilingual}, ReFinED \citep{ayoola-etal-2022-improving}, ExtEnD \citep{barba-etal-2022-extend}, CLOCQ\footnote{Although CLOCQ is designed to output multiple candidate entities for each mention, for the sake of our evaluation we compare to the target only its top-ranked one.} \citep{Christmann2022clocq}, and ReLiK \citep{Orlando2024}, as discussed in Section \ref{sec:related}. Each model has been used following the instructions provided by the authors in their respective websites or repositories. \textcolor{orange}{Additionally, we experiment with Large Language Models (LLMs) and test the ability of \textcolor{orange}{GPT-4 o1-mini}\footnote{\url{https://openai.com/index/gpt-4o-mini-advancing-cost-efficient-intelligence/}, the model recommended by OpenAI as the optimal choice for cost-efficiency trade-off as of December 2024.} and \textcolor{orange}{LLAMA 3.3 70B} instruction tuned\footnote{\url{https://www.llama.com/docs/model-cards-and-prompt-formats/llama3_3/}}. For both models, we rely on the same prompt instructing the model to produce the name of the Wikipedia page of a given named entity mention, or to answer with \texttt{NIL} if there is no relevant page. We report the prompt used in the appendix at section \ref{app:prompt}.}

For \modelCBLINK{} and \modelEL{}, as anticipated in Section \ref{sec:models}, we implement two plausibility functions addressing time and type constraints, detailed in Sections \ref{sec:time-c} and \ref{sec:type-c} respectively. Additionally, we explore various approaches for \texttt{NIL} prediction within \modelCBLINK{}, employing both heuristic methods and machine learning classifiers, which are elaborated in Section \ref{sec:nil-h}.

\subsection{Time constraint} \label{sec:time-c}
The first criterion on which we base our candidate plausibility function is chronological consistency (time plausibility).
We define $\phi_d \in \Phi$, the function that implements the time plausibility constraint such that $\phi_d(a, b) = 1$ if and only if $b$ chronologically follows $a$. In this case, the date of $a$ is the document's date of publication, which is given as metadata in HIPE-2020 and \benchmarkEL{}, and the date of $b$ is retrieved from Wikidata. We consider a wide set of time-related Wikidata properties reported in Table \ref{tab:timerelated_wd_prop} in the Appendix. The properties' retrieval order has been set according to a qualitative analysis of the semantics of the properties' names and descriptions. The analysis aimed to ensure to retain, when available, the most precise properties' values, such as \textsc{P569 - date of birth} (ranked first), against less precise ones, such as \textsc{P585 - Point in time} (ranked last).

\subsection{Type constraint} \label{sec:type-c}
The second criterion for candidate plausibility is comparing the types of two entities. Intuitively, if the mention is annotated as a human, entities of incompatible type e.g. building should be discarded.

We define $\phi_t \in \Phi$, the function that implements the type plausibility constraint such that $\phi_t(a, b) = 1$ if and only if the types of $a$ and $b$ are compatible. The types of $a$, i.e. the type of the mention, is the gold NER class assigned to the entity mention by the human annotators (for details about the types used in \benchmarkEL{}, see Appendix \ref{app:A}; for details about the types used in \textsc{HIPE}-2020, see \citet{HIPE2022ExtOverview}). The types of $b$ are extracted by retrieving the value of the \textsc{P31 - 'Instance of'} Wikidata property and are manually mapped to NER classes used in \benchmarkEL{} and HIPE-2020 by the same annotators that performed the annotation (see Section \ref{subsub:ann_guid}).
Compatibility between NER classes used in \benchmarkEL{} and HIPE-2020 and Wikidata types is further refined by compiling an \textit{ad-hoc} taxonomy (see Appendix \ref{app:A}) and checking whether the intersection between the types of $a$ and $b$ is not empty.

\subsection{NIL heuristics} \label{sec:nil-h}
As reported in Section \ref{sub:c-blink}, we implement the possibility of predicting \texttt{NIL} in \modelCBLINK{} by assuming that if no candidate has a notably higher similarity with the mention than the others, a confident link is absent. In our experiments, we extensively test all \texttt{NIL} heuristics listed in Table \ref{tab:nil_heuristics} to assess their impact on the output. For threshold-based heuristics, we experiment with values $\tau \in [0, 1]$ with increments of $0.001$. For machine learning approaches, we train classifiers on HIPE-2020 and evaluate their performance on \benchmarkEL{}.

\subsection{Results} \label{sec:results}

This section reports the results of the experiments conducted with \modelEL{}, \modelCBLINK{} and the considered SotA models on HIPE-2020 and \benchmarkEL{}. We report each constraint's influence and the heuristics' influence on the systems' accuracy. We evaluate our methods using the same evaluation metrics introduced by \citet{HIPE2022ExtOverview}. We follow the same experimental setting for all models: we consider a \texttt{NIL} annotation only if the model natively (or by extension, as in \modelCBLINK{}) supports \texttt{NIL} predictions. If the model does not provide any answer to our sentence, we consider it as wrongly linked. Table \ref{tab:MHERCL_results} reports the results for all the experiments on both \benchmarkEL{} and HIPE-2020. Different observations emerge from Table \ref{tab:MHERCL_results}, which are reported in the following paragraphs.

\paragraph{Off-the-shelf SotA entity linkers and LLMs fall short on HEL.}

\begin{table}[h]
    \parbox{.65\linewidth}{
        \caption{Percentage of mentions presenting OCR errors over the total of correctly and incorrectly linked entities per model. Models with a high number of empty answers, as reported in Table \ref{tab:empty_candidates}, are excluded.}
        \centering
        \begin{tabular}{llcc}
            \toprule
            Model & Dataset & \multicolumn{2}{c}{OCR errors (\%)} \\ \cmidrule{3-4}
            & & Over correct & Over wrong \\
            \midrule
            \multirow[t]{2}{*}{GENRE} & HIPE-2020 & $0.11$ & $0.33 $\\
             & \benchmarkEL{} & $0.05$ & $0.19$ \\
            \cmidrule{2-4}
            \multirow[t]{2}{*}{mGENRE} & HIPE-2020 & $0.09$ & $0.34$\\
             & \benchmarkEL{} & $0.07$ & $0.19$ \\
            \cmidrule{2-4}
            \multirow[t]{2}{*}{ReFinED} & HIPE-2020 & $0.21$& $0.31$ \\
             & \benchmarkEL{} & $0.07$& $0.20$ \\
            \cmidrule{2-4}
            \multirow[t]{2}{*}{\textcolor{orange}{LLAMA 3.3 70B}} & HIPE-2020 & $0.24$& $0.30$ \\
             & \benchmarkEL{} & $0.11$& $0.17$ \\
            \cmidrule{2-4}
            \multirow[t]{2}{*}{\textcolor{orange}{GPT-4 o1-mini}} & HIPE-2020 & $0.26$& $0.28$ \\
             & \benchmarkEL{} & $0.11$& $0.17$ \\
            \cmidrule{2-4}            
            \cmidrule{2-4}
            \multirow[t]{2}{*}{BLINK$^\dagger$} & HIPE-2020 & $0.09$ & $0.36$\\
             & \benchmarkEL{} & $0.08$ & $0.18$\\
            \cmidrule{2-4}
            \multirow[t]{2}{*}{BLINK} & HIPE-2020 & $0.10$& $0.35$ \\
             & \benchmarkEL{} & $0.09$ & $0.19$ \\
            \cmidrule{2-4}
    
            \multirow[t]{2}{*}{\modelCBLINK{}$^\star$} & HIPE-2020 & $0.19$ & $0.43$\\
             & \benchmarkEL{} & $0.10$ & $0.22$ \\
            \cmidrule{2-4}
            \multirow[t]{2}{*}{ELD$^\star$} & HIPE-2020 & $0.20$ & $0.39$ \\
            & \benchmarkEL{} & $0.09$ & $0.20$ \\
            \bottomrule
        \end{tabular}
        \label{tab:ocr_errors_impact}
    }
    \hfill
    \parbox{.3\linewidth}{
        \centering
        \caption{Percentages of no responses (no-candidates output) for the models that frequently fail to provide any linkable candidates for the mentions, leading to invalid linking and poor performance. Models from Table \ref{tab:MHERCL_results} for which such a scenario does not occur are excluded.}
        \begin{tabular}{ll|c}
            \toprule
            Model & Dataset & $\%$ \\
            \midrule    
            \multirow[t]{2}{*}{REL} & HIPE-2020 & $0.70$ \\
            & \benchmarkEL{} & $0.84$ \\
            \multirow[t]{2}{*}{CLOCQ} & HIPE-2020 & $0.80$ \\
            & \benchmarkEL{} & $0.71$ \\
            \multirow[t]{2}{*}{ExtEnD} & HIPE-2020 & $0.95$ \\
             & \benchmarkEL{} & $0.66$ \\
            \multirow[t]{2}{*}{ReLiK} & HIPE-2020 & $0.93$ \\
             & \benchmarkEL{} & $0.64$ \\
            \bottomrule
        \end{tabular}
        \label{tab:empty_candidates}
    }
\end{table}

\begin{table}[ht]
    \caption{Influence of entity popularity on EL performance. For \modelEL{} and \modelCBLINK{}, we consider their best runs as indicated in Table \ref{tab:MHERCL_results}. Spearman correlations marked with $^*$ denote statistical significance ($p < 0.01$) between popularity and correctness of the prediction. Models with a high number of no responses, as reported in Table \ref{tab:empty_candidates}, are excluded from this analysis. The lowest \% increase and correlation coefficient are underlined, while the highest are marked in bold.}
    \centering
    \begin{tabular}{ll|l|l}
        \toprule
        Model & Dataset & More popular QID chosen (\%) & Spearman correlation \\
        \midrule
        \multirow[t]{2}{*}{GENRE} & HIPE-2020 & $\mathbf{0.25}$ & $\mathbf{0.52}^*$ \\
         & \benchmarkEL{} & $\mathbf{0.16}$ & $\mathbf{0.55}$$^*$ \\
        \cmidrule{2-4}
        \multirow[t]{2}{*}{mGENRE} & HIPE-2020 & $0.18$ & $0.40$$^*$ \\
         & \benchmarkEL{} & $0.14$ & $0.49$$^*$ \\
        \cmidrule{2-4}
        \multirow[t]{2}{*}{ReFinED} & HIPE-2020 & $\underline{0.07}$ & $0.17$ \\
         & \benchmarkEL{} & $\underline{0.05}$ & $0.28$$^*$ \\
        \cmidrule{2-4}       
        \cmidrule{2-4}
        \multirow[t]{2}{*}{BLINK$^\dagger$} & HIPE-2020 & $0.12$ & $0.40$$^*$ \\
         & \benchmarkEL{} & $0.11$ & $0.31$$^*$ \\
        \cmidrule{2-4}
        \multirow[t]{2}{*}{BLINK} & HIPE-2020 & $0.14$ & $0.40$$^*$ \\
         & \benchmarkEL{} & $0.10$ & $0.33$$^*$ \\
        \cmidrule{2-4}
        \multirow[t]{2}{*}{\textcolor{orange}{LLAMA 3.3 70B}} & HIPE-2020 & $0.10$ & $0.32$$^*$  \\
         & \benchmarkEL{} & $0.05$ & $0.21$$^*$ \\
         \cmidrule{2-4}
        \multirow[t]{2}{*}{\textcolor{orange}{GPT-4 o1-mini}} & HIPE-2020 & $0.08$ & $0.31$$^*$ \\
         & \benchmarkEL{} & $0.06$ & $0.39$$^*$ \\         
        \midrule        
        \multirow[t]{2}{*}{\modelCBLINK{}$^\dagger_{\{ \phi_d, \phi_t \}}$} & HIPE-2020 & $0.14$ & $0.40$$^*$ \\
        & \benchmarkEL{} & $0.08$ & $0.34$$^*$ \\
        \cmidrule{2-4}
        \multirow[t]{2}{*}{\modelCBLINK{}$_{\{ \phi_d, \phi_t \}}$} & HIPE-2020 & $0.14$ & $0.38$$^*$ \\
         & \benchmarkEL{} & $0.10$ & $0.31$$^*$ \\
        \cmidrule{2-4}
        \multirow[t]{2}{*}{\modelCBLINK{}$^\star$} & HIPE-2020 & $0.09$ & $0.28$$^*$ \\
         & \benchmarkEL{} & \underline{$0.03$} & $0.22$$^*$ \\
        \cmidrule{2-4}
        \multirow[t]{2}{*}{ELD$^\dagger$} & HIPE-2020 & $\underline{0.07}$ & $\underline{-0.24}^*$ \\
         & \benchmarkEL{} & $0.06$ & $0.18$$^*$ \\
         \cmidrule{2-4}
        \multirow[t]{2}{*}{ELD$^\dagger_{\{ \phi_d, \phi_t \}}$} & HIPE-2020 & $\underline{0.07}$ & $-0.28$$^*$ \\
         & \benchmarkEL{} & $0.06$ & $\underline{0.05}$ \\
         \cmidrule{2-4}
        \multirow[t]{2}{*}{ELD} & HIPE-2020 & $0.18$ & $0.41$$^*$ \\
         & \benchmarkEL{} & $\textbf{0.16}$ & $0.48$$^*$ \\
         \cmidrule{2-4}

        \multirow[t]{2}{*}{ELD$^\star$} & HIPE-2020 & $0.16$ & $0.40$$^*$ \\
         & \benchmarkEL{} & $0.10$ & $0.48$$^*$ \\
        \bottomrule
    \end{tabular}
    \label{tab:morepop_qid_preferred}
\end{table}

Table \ref{tab:MHERCL_results} shows that SotA \textcolor{orange}{specialised} EL models significantly underperform in historical document contexts, frequently failing to produce linkable entities for most sentences in both datasets. In particular, REL, CLOCQ, ExtEnD and ReLiK do not produce any linking candidate for mentions in the majority of the cases, as reported in Table \ref{tab:empty_candidates}. This issue underscores a critical limitation in current approaches when faced with the specific challenges of historical texts, such as OCR errors and differences in morphosyntactic and semantic features between contemporary and historical language.

OCR errors significantly impact the performance of EL models. Table \ref{tab:ocr_errors_impact} presents the proportion of mentions affected by OCR errors over correctly and incorrectly linked mentions. OCR errors are more frequent in mentions linked incorrectly than in mentions linked correctly for all tested models. These errors are uncommon in typical NLP tasks and infrequent in pre-training corpora, leading PLMs to produce suboptimal representations. However, they are frequently found in historical documents (Table \ref{tab:MHERCLv01_HIPE2020_NamedEntitiesStats}), making datasets like HIPE-2020 and \benchmarkEL{} valuable for evaluating NLP robustness to OCR errors. \textcolor{orange}{As observed in Table \ref{tab:ocr_errors_impact}, the proportions of mentions with OCR errors between correctly and incorrectly linked mentions are less unequal for the two LLMs tested (\textcolor{orange}{LLAMA 3.3 70B} and GPT-4-o1-mini) compared to the other models. This suggests that recent LLMs with an high number of parameters are more robust to OCR errors than models specifically designed for the EL task, and this factor may account for their superior performance compared to all other SotA specialised models tested.}

Moreover, \textcolor{orange}{specialised} models jointly performing mentions recognition and linking generally underperform compared to the others. For instance, ReLiK and ExtEnD, which approach the EL task as an extraction problem, yield worse results than models requiring the gold NER information to be provided explicitly in their input, such as GENRE, mGENRE and BLINK. This underlines that historical datasets are also a valuable asset for NER. In particular, the high number of unique named entities in \benchmarkEL{} makes it a convenient benchmark for assessing NER accuracy on long-tail entities.
Indeed, we highlight that ReFinEd employs a similar architecture but achieves competitive results comparable to models explicitly receiving the gold NER information as input. This is because it is designed to utilize a broad spectrum of symbolic KB information that enhances its accuracy with long-tail and ambiguous entities.

GENRE and mGENRE require a prior named entity recognition phase and generally perform better than extraction-based models. However, they still underperform when compared to retrieval-based models such as BLINK. This trend is also supported by \modelEL{}'s results, which operates similarly to a retrieval-based model. Despite their higher efficiency, generative-based models are unable to consider a large pool of linkable candidates since they must only consider a small subset of possible entities to maintain a tractable linking procedure.

Given these considerations, we argue that retrieval-based methods are less prone to popularity bias, as their encoding and retrieval phases can be designed to account for it. For instance, the performance of \modelEL{} relative \textcolor{orange}{to the other SotA specialised models tested} indicates that it is less sensitive to popularity bias. Typically, \textcolor{orange}{SotA specialised models} fine-tune the PLM on datasets suffering from popularity bias, as discussed in Section \ref{sec:popularity}. If the PLM natively suffers from sub-optimal representations on long-tail entities, the fine-tuning approach will reinforce this aspect, further degrading its performances on historical documents. Since \modelEL{} relies on a PLM that is not specifically fine-tuned to accommodate the EL retrieval phase, it can rely on less biased entity representations.


\textcolor{orange}{The LLMs tested (LLAMA 3.3 70B and GPT-4 o1-mini) generally outperform the SotA specialised models in terms of accuracy. Their extensive computational resources and large-scale training data enable robustness to OCR noise and effective contextualization of sentence-level information based on the data encountered during training. Specifically, LLAMA 3.3 70B, which provides access to the model but does not disclose its training dataset, achieves superior performance compared to all specialised SotA EL models tested on \benchmarkEL{}. Although the computational effort required for this model is substantially higher than that of the other baselines, its high accuracy indicates that a distillation-based approach \citep{gu2024llmdistill} could enhance retrieval-based systems by improving their retrieval components. Conversely, GPT-4 o1-mini demonstrates superior performance across all models on HIPE-2020 and achieves results comparable to LLAMA 3.3 70B on \benchmarkEL{}. However, the lack of openly available model weights for GPT-4 limits its flexibility for integration with other approaches. Furthermore, caution is required regarding potential data contamination issues, particularly for the HIPE-2020 dataset, which has been available since 2020. This concern aligns with observations raised in recent studies by \citet{jacovi-etal-2023-stop}, \citet{sainz-etal-2023-nlp}, and \citet{Li_Flanigan_2024}.}

Table \ref{tab:morepop_qid_preferred} quantitatively examines how entity popularity affects EL model performance, reporting the frequency with which more popular QIDs are chosen over less popular ones and the Spearman correlation between prediction correctness and entity popularity. \modelCBLINK{}$^\star$, \modelEL{}$^\dagger$, and ReFinED exhibit a lower tendency to favour popular QIDs on both \benchmarkEL{} and HIPE-2020 datasets. For \modelCBLINK{}$^\star$ and \modelEL{}$^\dagger$, this reduced bias stems from the filtering approach discussed in Section \ref{sec:time-c} and \ref{sec:type-c}, which excludes implausible entities based on temporal or type-related constraints, minimizing the chances of incorrectly selecting a more popular entity when it is logically implausible. ReFinED, designed specifically for robust performance with long-tail entities, demonstrates the lowest preference for more popular QIDs, as reflected in its Spearman correlation, which is remarkably lower than other models across both datasets. Conversely, the generative-based models, GENRE and mGENRE, show a stronger tendency to favour popular entities, reflecting a greater bias toward high-frequency entities.

In contrast, \modelEL{}'s unsupervised approach relies on a sentence representation which is agnostic to the final task, eventually ensuring a fairer comparison between entities, which results in higher performance in HEL. Indeed, the results of \modelEL{}$^\dagger$, which does not use replicator dynamics, have a low correlation with the entities' popularity. Note, however, that using \modelEL{} with the replicator dynamics results in better quantitative performances (Table \ref{sec:results}) but displays a positive correlation with popularity bias. This depends on the PLM employed during the encoding phase. If the PLM produces representations that ``mildly'' favour popular entities, the replicator dynamics process is akin to an online fine-tuning process, reinforcing the bias. It follows that there is a trade-off between quantitative performances and popularity bias using this approach. Further attention must be devoted to producing entity representations that reduce or eliminate bias towards popular entities.

\textcolor{orange}{The LLMs tested (LLAMA 3.3 70B and GPT-4 o1-mini) appear more robust to popularity variations, as evidenced by their percentage of more popular QIDs chosen over less popular ones and Spearman correlations comparable to those of \modelCBLINK{}$^\star$, \modelEL{}$^\dagger$, which are designed to logically exclude implausible entities, and ReFinED, which is designed to achieve better performance on long-tail knowledge scenarios. Their massive-scale training data and architectural complexity make them competitive in linking less popular entities.}

Interestingly, all the \textcolor{orange}{specialised SotA EL models} perform better on \benchmarkEL{} than on HIPE-2020, despite \benchmarkEL{} being composed of less popular entities as compared to HIPE-2020 (see Section \ref{sec:popularity}). This discrepancy is evident in BLINK, which, while underperforming on HIPE-2020, emerges as the best-performing baseline on \benchmarkEL{}. The difference in performance can be attributed to the higher incidence of \texttt{NIL} entities in HIPE-2020 compared to \benchmarkEL{}. Employing a conservative strategy that invariably predicts \texttt{NIL} already establishes a robust baseline. In contrast, the lower prevalence of \texttt{NIL} entities in \benchmarkEL{} facilitates more effective testing of SotA entity linkers on historical documents, providing a clearer assessment of their capabilities.

\paragraph{Time and type constraints enable performance improvement on HEL.}

\begin{table}[ht]
    \caption{Accuracy and F1 score computed on the type and time plausibility of SotA models and our proposed models' predictions. Models with a high number of no responses, as reported in Table \ref{tab:empty_candidates}, are excluded from this analysis. Subscripts indicate the metric. An empty answer is considered wrong. Mentions labelled with \texttt{NIL} are excluded. The best results are represented in bold. The worst results are underlined.}
    \centering
    \begin{tabular}{ll|rr|rr}
        \toprule
        Model & Dataset & Year$_A$ & Year$_{F1}$ & Type$_A$ & Type$_{F1}$ \\
        \midrule

        \multirow[t]{2}{*}{GENRE} & HIPE-2020 & 0.67 & 0.80 & 0.60 & 0.75 \\
         & \benchmarkEL{} & 0.70 & 0.82 & 0.77 & 0.87 \\
        \cmidrule{2-6}
        \multirow[t]{2}{*}{mGENRE} & HIPE-2020 & 0.63 & 0.77 & 0.54 & 0.70 \\
         & \benchmarkEL{} & 0.76 & 0.87 & 0.77 & 0.87 \\
        \cmidrule{2-6}
        \multirow[t]{2}{*}{ReFinED} & HIPE-2020 & \underline{0.27} & \underline{0.43} & \underline{0.25} & \underline{0.40} \\
         & \benchmarkEL{} & 0.55 & 0.71 & 0.54 & 0.70 \\
        \cmidrule{2-6}
        \cmidrule{2-6}
        \multirow[t]{2}{*}{BLINK} & HIPE-2020 & 0.74 & 0.85 & 0.66 & 0.80 \\
         & \benchmarkEL{} & 0.85 & 0.92 & 0.85 & \textbf{0.92} \\
        \cmidrule{2-6}
        \multirow[t]{2}{*}{BLINK$^\dagger$} & HIPE-2020 & 0.70 & 0.82 & 0.67 & 0.80 \\
         & \benchmarkEL{} & 0.79 & 0.88 & 0.80 & 0.89 \\ 
         \cmidrule{2-6}
        \multirow[t]{2}{*}{\textcolor{orange}{LLAMA 3.3 70B}} & HIPE-2020 & \textbf{0.79} & \textbf{0.89} & \textbf{0.68} & \textbf{0.81} \\
        & \benchmarkEL{} & 0.80 & 0.89 & 0.74 & 0.85 \\
        \cmidrule{2-6}
        \multirow[t]{2}{*}{\textcolor{orange}{GPT-4 o1-mini}} & HIPE-2020 & 0.72 & 0.84 & 0.60 & 0.75 \\
        & \benchmarkEL{} & 0.63 & 0.77 & 0.61 & 0.76 \\
        \midrule

        \multirow[t]{2}{*}{\modelCBLINK{}$^\star$} & HIPE-2020 & 0.74 & 0.85 & 0.66 & 0.80 \\
         & \benchmarkEL{} & \textbf{0.87} & \textbf{0.93} & \textbf{0.86} & \textbf{0.92} \\
        \cmidrule{2-6}
        \multirow[t]{2}{*}{ELD$^\star$} & HIPE-2020 & 0.51 & 0.68 & 0.44 & 0.61 \\
        & \benchmarkEL{} & \underline{0.51} & \underline{0.68} & \underline{0.50} & \underline{0.66} \\
        \bottomrule
    \end{tabular}
    \label{tab:confusion_metrics_time_type_filters}
\end{table}

The results from Table \ref{tab:MHERCL_results} prove that plausibility filters (as introduced in Sections \ref{sec:time-c} and \ref{sec:type-c}) are a straightforward way of enhancing EL methods when historical documents are considered. Both \modelEL{} and \modelCBLINK{} demonstrate equal or superior performance when filters are enabled.


 \newcommand{\mcml}[1]{\multicolumn{5}{p{\dimexpr\linewidth-2\tabcolsep-2\arrayrulewidth}}{\footnotesize #1}}

\begin{table}[!htbp]
    \centering
    \caption{Examples of predictions made on \benchmarkEL{} and HIPE-2020. The target mention is underlined, and its QID, Wikipedia page title, NEC class and Wikidata year are in brackets. We report the predicted QID, Wikipedia page title, and Wikidata type and time information. Matching information is in bold. \modelCBLINK{}$^\star$ and \modelEL{}$^\star$ are the best models in Table \ref{tab:MHERCL_results}. Models with a high number of no responses (Table \ref{tab:empty_candidates}) are excluded.}
    \label{tab:MHERCL_results_example}

    \footnotesize
    \begin{tabularx}{\textwidth}{ccXXc}
    \toprule

    \multicolumn{5}{c}{The Quarterly Musical Magazine And Review, 1825 (\benchmarkEL{})} \\
    \mcml{\textit{\underline{M. Barriere} (\texttt{Q726908, Jean-Baptiste Barrière, person, 1707}}) has published four sets of quartets, and several Symphonies, concertos, trios, and duets.} \\ \midrule
    Model & Prediction & Wikipedia title & Type & Year \\ \midrule
    BLINK & Q3086402 & Françoise Barrière & \textbf{human} & 1944 \\
    GENRE & Q56537639 & Barrière & family name & \\
    mGENRE & Q3588326 & Émile Barrière & \textbf{human} & 1902\\
    ReFinED & \texttt{NIL} & & & \\
    \textcolor{orange}{LLAMA 3.3 70B} & \textbf{Q726908} & \textbf{Jean-Baptiste Barrière} & \textbf{human} & \textbf{1707} \\
    \textcolor{orange}{GPT-4 o1-mini} & \texttt{NIL} & & & \\
    \modelCBLINK{}$^\star$ & \textbf{Q726908} & \textbf{Jean-Baptiste Barrière} & \textbf{human} & \textbf{1707} \\
    \modelEL{}$^\star$ & \texttt{NIL} & & & \\ \midrule

    \multicolumn{5}{c}{The Musical Times, 1873 (\benchmarkEL{})} \\
    \mcml{\textit{On the 14th ult the South Wales Choral Union visited \underline{Marlborough Honse} (\texttt{Q565532, Marlborough House, building, 1711}), by express desire of His Royal Highness the Prince of Wales.}} \\ \midrule
    Model & Prediction & Wikipedia title & Type & Year \\ \midrule
    BLINK & Q539528 & Marlborough, Wiltshire & market town & \\
    GENRE & Q539528 & Marlborough, Wiltshire & market town & \\
    mGENRE & Q1264586 & Duke of Marlborough (title) & noble title & \\
    ReFinED & \texttt{N/A} &  &  & \\
    \textcolor{orange}{LLAMA 3.3 70B} & \textbf{Q565532} & \textbf{Marlborough House} & \textbf{mansion} & \textbf{1711}\\
    \textcolor{orange}{GPT-4 o1-mini} & \texttt{NIL} & & & \\
    \modelCBLINK{}$^\star$ & \textbf{Q565532} & \textbf{Marlborough House} & \textbf{mansion} & \textbf{1711}\\
    \modelEL{}$^\star$ & \texttt{NIL} & & & \\
    \midrule

    \multicolumn{5}{c}{\textcolor{orange}{The Quarterly Musical Magazine and Review, 1828 (\benchmarkEL{})}} \\
    \multicolumn{5}{l}{\textcolor{orange}{\textit{\underline{Furno} (\texttt{Q555808, Giovanni Furno, person, 1748}), who fills this post, is equally active and skilful.}}} \\
    \midrule
    \textcolor{orange}{Model} & \textcolor{orange}{Prediction} & \textcolor{orange}{Wikipedia title} & \textcolor{orange}{Type} & \textcolor{orange}{Year} \\
    \midrule
    \textcolor{orange}{BLINK} & \textcolor{orange}{Q920465} & \textcolor{orange}{Carlo Furno} & \textcolor{orange}{\textbf{human}} & \textcolor{orange}{1921} \\
    \textcolor{orange}{GENRE} & \textcolor{orange}{Q1475080} & \textcolor{orange}{Furno} & \textcolor{orange}{Wikimedia Disambiguation Page} & \textcolor{orange}{} \\
    \textcolor{orange}{mGENRE} & \textcolor{orange}{Q1475080} & \textcolor{orange}{Furno} & \textcolor{orange}{Wikimedia Disambiguation Page} & \textcolor{orange}{} \\
    \textcolor{orange}{ReFinED} & \textcolor{orange}{Q104052585} & \textcolor{orange}{Furno} & \textcolor{orange}{\textbf{human}} & \textcolor{orange}{} \\
    \textcolor{orange}{LLAMA 3.3 70B} & \textcolor{orange}{Q517279} & \textcolor{orange}{Antonio Maria Vegliò} & \textcolor{orange}{\textbf{human}} & \textcolor{orange}{1938} \\
    \textcolor{orange}{GPT-4 o1-mini} & \textcolor{orange}{\texttt{NIL}} & \textcolor{orange}{} & \textcolor{orange}{} & \textcolor{orange}{} \\
    \textcolor{orange}{\modelCBLINK{}$^\star$} & \textcolor{orange}{\textbf{Q555808}} & \textcolor{orange}{\textbf{Giovanni Furno}} & \textcolor{orange}{\textbf{human}} & \textcolor{orange}{\textbf{1748}} \\
    \textcolor{orange}{\modelEL{}$^\star$} & \textcolor{orange}{\texttt{NIL}} & \textcolor{orange}{} & \textcolor{orange}{} & \textcolor{orange}{} \\
    \midrule

    \multicolumn{5}{c}{\texttt{sn83030483}, 1790 (HIPE-2020)} \\
    \multicolumn{5}{c}{\textit{\underline{JOHN SULLIVAN.} (\texttt{Q886420, John Sullivan, person, 1740}).}} \\ \midrule
    Model & Prediction & Wikipedia title & Type & Year \\ \midrule
    BLINK & \texttt{Q6259624} & John Sullivan (VC) & \textbf{human} & 1830 \\
    GENRE & \texttt{Q6259628} & John Sullivan (pitcher) & \textbf{human} & 1894 \\
    mGENRE & \texttt{Q11082552} & John Sullivan & Wikimedia human name disambiguation page & \\
    ReFinED & \texttt{Q3396287} & John Sullivan (writer) & \textbf{human}& 1946\\
    \textcolor{orange}{LLAMA 3.3 70B} & \texttt{Q11082552} & John Sullivan & Wikimedia human name disambiguation page & \\
    \textcolor{orange}{GPT-4 o1-mini} & \texttt{Q11082552} & John Sullivan & Wikimedia human name disambiguation page & \\
    \modelCBLINK{}$^\star$ & \texttt{NIL} & & & \\
    \modelEL{}$^\star$ & \textbf{Q886420} & \textbf{John Sullivan (general)} & \textbf{human} & \textbf{1740} \\ \bottomrule
    \end{tabularx}
\end{table}

For instance, \modelCBLINK{} with time and type constraints can link historical entities where most the other models fail. In the first example of Table \ref{tab:MHERCL_results_example}, all the tested models but \modelCBLINK{}$^\star$ \textcolor{orange}{and LLAMA 3.3 70B} produce links to entities that are implausible because born after the date of issue of the historical periodical from which the sentence was extrapolated. Note that ReFinED, \modelEL{}$^\star$ \textcolor{orange}{and GPT-4 o1-mini} are overly conservative in predicting a NIL link. Similarly, in the second example of Table \ref{tab:MHERCL_results_example}, all the tested models but \modelCBLINK{}$^\star$ \textcolor{orange}{and LLAMA 3.3 70B} produce inconsistent links regarding the entity type. \textcolor{orange}{\modelCBLINK{}$^\star$ leverages the implemented plausibility filters to correctly predict the expected entities}. Such a successful linking happens despite the OCR error in the mention, demonstrating that plausibility filters can also be helpful in cases where the mention presents OCR noise. Once again, ReFinED and \modelEL{}$^\star$ conservatively predict NIL. \textcolor{orange}{Interestingly, LLAMA 3.3 70B can overcome the OCR errors affecting the mention and leverages the context of the sentence to correctly identify the expected mention, while GPT-4 o1-mini does not show the same behaviour and rather predicts \texttt{NIL}. Multiple reasons can explain this aspect, given the opaque inference mechanism that characterizes LLMs. We posit that the different instruction tuning processes employed in specializing both models result in a model, GPT-4 o1-mini, being more conservative in its predictions. This also explains the higher performances of this model in Table \ref{tab:MHERCL_results}, where models that consistently rely on \texttt{NIL} links outperform the other models.} \textcolor{orange}{In the third example, \modelCBLINK{}$^\star$ is the only model able to predict the correct entity. This success is attributed to its plausibility filters, which restrict the model to considering only entities that are valid in terms of both type and time information. In contrast, the other models fail by either suggesting an entity with an incompatible type or one with an implausible date.}
In the \textcolor{orange}{final} example, instead, \modelEL{}$^\star$ is the only model that predicts the correct entity thanks to the application of the plausibility filters, while \modelCBLINK{}$^\star$ is excessively conservative and predicts a NIL link. In this case, the other models do not find the correct entity, either because of type or date implausibility.

In Table \ref{tab:confusion_metrics_time_type_filters}, we report the accuracy and F1-score for each model in predicting a plausible entity by framing it as a classification task. A predicted entity is considered plausible with respect to the target entity if its time information, retrieved as discussed in Section \ref{sec:time-c}, precedes the issue date of the source document containing the sentence where the target entity appears. Similarly, it is considered plausible if its type matches that of the target entity. BLINK, GENRE, mGENRE \textcolor{orange}{and LLAMA 3.3 70B} obtain good performances on both time and type plausibility, even when compared to \modelCBLINK{}. This provides evidence that when relying on a large amount of training data, those models can infer some level of plausibility. \textcolor{orange}{In particular, LLAMA 3.3 70B outperforms all other models in plausibility accuracy on HIPE-2020, even the ones including filters. Nonetheless, its performances in Table \ref{tab:MHERCL_results} are significantly lower than the ones of GPT-4 o1-mini. This is explained by the tendency of GPT-4 o1-mini to favour \texttt{NIL} predcitions, as seen in the examples of Table \ref{tab:MHERCL_results_example}, and underscores the importance of this aspect in HEL. We highlight that the plausibility performances of LLAMA 3.3 70B are not consistent between HIPE-2020 and \benchmarkEL{}. This further reinforces the argument that \benchmarkEL{} is a valuable dataset in the NLP landscape.}

Note that the application of time and type constraints on \modelCBLINK{} and \modelEL{} do not always restrict the pool of candidates to only those that are plausible. This is due to the mapping between NERC types and Wikidata types that we designed. Nonetheless, \modelCBLINK{} demonstrates a higher ability to propose a plausible candidate when compared to BLINK.

We remark that plausibility constraints are crucial for enabling the application of the \texttt{NIL} heuristics discussed in Section \ref{sec:nil-h}. 
These heuristics assume that all candidates considered by the model are plausible options. In cases where no plausible candidate has a significantly higher similarity score than the others, we assume that the model cannot confidently establish a reliable link and should instead output \texttt{NIL}. This heuristic would be impossible to apply without a plausibility assumption since implausible entities might still display a high similarity score due to the PLMs tendency to favour similarity with popular entities.

\paragraph{Effective models for HEL must take \texttt{NIL} into account.}
We experiment with several \texttt{NIL} heuristics and observe different outcomes. In general, machine learning-based heuristics fail to generalize to different distributions, as seen in \modelCBLINK{} relying on decision trees. The best approach is \modelCBLINK{}$^\dagger$ with the deviation from the mean heuristics. By avoiding the re-ranking phase, the scores produced by BLINK$^\dagger$ during the candidate scoring phase are more reliable in \texttt{NIL} prediction. Since we remove every implausible candidate, the model can conservatively avoid a wrong link if no remaining candidate significantly stands out. \texttt{NIL} heuristics based on the similarity between superficial mentions are less effective than expected. This can be attributed to OCR errors and the different language registries employed by historical documents compared to the one found in the reference KB.

Our initial hypothesis on the importance of managing \texttt{NIL} predictions is empirically confirmed by the fact that the best \texttt{NIL} heuristic for \modelCBLINK{} largely outperforms all the other models on both HIPE-2020 and \benchmarkEL{}. \textcolor{orange}{Despite allowing tested LLMs to link named entity mentions to \texttt{NIL} when no suitable Wikipedia page title could be suggested, the proposed \modelCBLINK{}$^\dagger$ with the deviation from the mean heuristic achieves the highest F1 score. The results highlight that a simpler retrieval-based neural entity linking model enhanced with targeted heuristics can outperform larger, more expensive, and closed-source models in long-tail, domain-specific scenarios requiring robust \texttt{NIL} handling.}

\textcolor{orange}{Additionally, it is notable that \modelEL{} using replicator dynamics, with type- and time-plausibility candidates filtering logic activated and \texttt{NIL} in candidates outperforms LLAMA 3.3 70B on HIPE-2020. On MHERCL, it still performs competitively, falling short of LLAMA 3.3 70B's performance by only $0.03$ F1 and GPT-4 o1-mini's by just $0.02$ F1.}


\begin{table}[ht]
\centering
\caption{Error analysis on \benchmarkEL{} for the best runs of \modelCBLINK{} and \modelEL{}. The \textit{Target} is the gold annotation, which can be \texttt{NIL} or a QID. The target can or cannot be among the candidates retrieved by the models. Note that \texttt{NIL} can never be among \modelCBLINK{} candidates, but it can be retrieved by \modelEL{}. The highest error type percentages per each model are indicated in bold text, while the lower ones are underlined.}
\begin{tabular}{cccccll}
    \toprule
    \multirow{2}{*}{Model} & \multirow{2}{*}{Target} & \multirow{2}{*}{\vtop{\hbox{\strut Is target}\hbox{in top-k candidates?}}} & \multirow{2}{*}{Wrong prediction} & \multicolumn{2}{c}{Errors (\%)} \\
    \cmidrule(lr){5-6}
    &  &  &  & HIPE-2020 & \benchmarkEL{} \\
    \midrule

    \multirow{5}{*}{\modelCBLINK{}$^\star$} & \texttt{NIL} & $-$ & QID & 0.14 & $0.17$ \\ \cmidrule(lr){2-6}
    & \multirow{4}{*}{QID} & $\times$ & \texttt{NIL} & $\mathbf{0.47}$ & $0.31$ \\
    &  & $\times$  & QID & \underline{0.05} & \underline{0.06} \\ \cmidrule(lr){3-6}
    &  & $\checkmark$ & \texttt{NIL} & 0.26 & $\mathbf{0.38}$\\
    &  & $\checkmark$ & QID & 0.07 & $0.08$\\

    \midrule
    \multirow{8}{*}{\modelEL{}$^\star$} & \multirow{2}{*}{NIL}& $\times$ & QID & 0.09 & $0.07$\\ \cmidrule(lr){3-6}
      &  & $\checkmark$ & QID & 0.05 & $0.06$\\  \cmidrule(lr){2-6}
     & \multirow{4}{*}{QID} & $\times$ & \texttt{NIL} & $\mathbf{0.64}$ & $\mathbf{0.64}$\\
     &  & $\times$ & QID & 0.11 & $0.18$ \\ \cmidrule(lr){3-6}
     &  & $\checkmark$ & \texttt{NIL} & \underline{0.0} & $\underline{0.002}$\\
     &  & $\checkmark$ & QID & 0.03 & $0.03$\\
    \bottomrule
\end{tabular}

\label{tab:err_analysis_mercl_clef}
\end{table}

Table \ref{tab:err_analysis_mercl_clef} provides an analysis of the incidence of \texttt{NIL} predictions for the best \modelCBLINK{} and \modelEL{} models in Table \ref{tab:MHERCL_results} on \benchmarkEL{} and HIPE-2020. The analysis is divided between mentions that should be linked to \texttt{NIL} and the others. Furthermore, whether the linking target is among the list of candidates at the disposal of the model is analyzed.

For \modelCBLINK{}, the most frequent error on \benchmarkEL{} occurs when the model predicts \texttt{NIL} despite the target QID being among the candidates. This is a consequence of our \texttt{NIL} prediction heuristics. Even though this approach allows better performances, they rely on the scores produced by the model and assume that their magnitude is a reliable proxy for their fitness as a final link. However, the retriever (and re-ranker) models are not explicitly trained for this task. It might hence happen that the target candidate is not perceived as significantly different from the other plausible results, hence resulting in the inference of \texttt{NIL}. Integrating a plausibility mechanism that can propagate the decision to the encoder is a promising approach that could result in more robust entity encoders that produce more reliable similarity scores.
The model's second most frequent error on \benchmarkEL{} and first most frequent error on HIPE-2020 happens when the model predicts \texttt{NIL} because the target QID is not within the candidates considered. Such a behavior can also be observed in \modelEL{}, where the highest \% of error types relate to cases where the target QID is not among the linking candidates, and the model predicts \texttt{NIL}. 
The least frequent errors for both models include cases where the target QID is absent, but the model predicts a wrong QID, or where the target QID is present, and the model selects a different one. In particular, \modelEL{} rarely prefers \texttt{NIL} over a correct QID and chooses a different (wrong) QID only in $3\%$ of the cases. As predicting \texttt{NIL} is optimal when the target QID is not in the candidate set, such a conservative approach is beneficial in scenarios in which precise prediction of QIDs is key, as in long-tail, domain-specific knowledge extraction scenarios. In such contexts, where low-popularity named entities are prevalent, it is often preferable to avoid incorrect linkages by opting for \texttt{NIL} rather than risking an inaccurate association. Thus, while the model's conservatism may lead to a higher rate of \texttt{NIL} predictions, it ultimately supports more precise outcomes in cases where the entities in question are less popular.

\section{Conclusion}
\label{sec:conclusion}
In this paper, we introduced \benchmarkEL{}, a gold standard benchmark of annotated English sentences for historical NER, NEC and EL. This dataset comprises sentences from music historical periodicals published between 1823 and 1900 and collected within the Polifonia Corpus\footref{foot:polifonia_corpus}. It is intended to support further study in HEL and, more generally, on the recognition and linking of long-tail and domain-specific named entities.

We also presented and evaluated \modelEL{}, an unsupervised model for EL, and \modelCBLINK{}, an extended version of BLINK incorporating plausibility filters that leverage knowledge from trusted KGs, in our case Wikidata. \modelEL{} demonstrates that an unsupervised model can be effective in HEL, addressing the popularity bias in training datasets by eliminating the need for a supervised training phase. Implementing \modelEL{} using replicator dynamics from game theory outperforms all baselines. The implementation of time and type constraints within \modelCBLINK{} has enabled a significant improvement of the model's performance on the tested historical documents benchmarks. Assuming that all candidates are plausible based on time and type allows \modelCBLINK{} to integrate heuristics predicting entities that are not in the KB of reference (\texttt{NIL} links), outperforming all the other tested models. When plausibility filters are applied, \modelEL{}'s performance improves further, underscoring the value of integrating these constraints in HEL tasks.

Future work will focus on expanding the benchmark to include annotated sentences in other languages, addressing a gap in resource availability for NER, NEC and EL tasks.
\textcolor{orange}{Moreover, we also aim at testing OCR-post correction methods \citep{DBLP:conf/aaai/Ramirez-OrtaXMM22} to try to limit the impact that OCR errors have on candidate generation and linking; and to complement OCR-post correction with data augmentation to increase lexical variability \citep{ijcai2021p528} \citep{lacerra-etal-2021-genesis} }.
Additionally, we plan to explore methods for integrating plausibility constraints into the training process of EL models and evaluate these approaches on standard EL benchmarks. \textcolor{orange}{Finally, the competitive accuracy of LLMs when prompted in a zero-shot fashion in the context of HEL is a promising approach, proving that their huge pre-training process helps in coupling better with the specifics of historical documents (OCR errors, morphosyntactic differences from contemporary language, etc.) as well as popularity bias}. In particular, the filtered retrieval approach employed for \modelCBLINK{} can be integrated into Retrieval Augmented Generation (RAG) methods, delegating to LLMs the final decision on the entity that is better suited as a link, \textcolor{orange}{exploiting their resilience to OCR errors}. \textcolor{orange}{Moreover, a particularly promising approach is to integrate an open-weights LLM, such as LLAMA 3.3 70B, with the approach used by (m)GENRE. Specifically, restricting the generation of candidates to those considered plausible would allow these models to enforce time and type plausibility, enhancing their capabilities and potentially improving their sensitivity to \texttt{NIL} predictions.}

\section{Declarations}
\subsection*{Funding}
This project has received funding from the European Union’s Horizon 2020 research and innovation programme under grant agreement No 101004746. Nicolas Lazzari is supported by the FAIR – Future Artificial Intelligence Research Foundation as part of the grant agreement MUR n. 341, code PE00000013 CUP 53C22003630006.

\subsection*{Competing interests}
The authors have no competing interests to declare that are relevant to the content of this article.

\subsection*{Data availability}
All the code for our experiments is publicly available on GitHub\footnote{ Available at the URL \url{https://github.com/polifonia-project/historical-entity-linking}.}. The \benchmarkEL{} dataset is available in the same GitHub repository and on HuggingFace\footnote{Available at the URL \url{https://huggingface.co/datasets/n28div/MHERCL}.}.

\bibliography{sn-bibliography}

\clearpage
\begin{appendices}
\section{Annotation Guidelines}
\label{app:B}

\paragraph*{General guidelines.} The annotators were instructed to consider a named entity a real-world thing indicating a unique individual through a proper noun \citep{JurafskyMartin2023}. The annotators were introduced to a typical NLP pipeline (tokenization, part-of-speech tagging, word sense disambiguation, NER, NEC and EL), emphasising the NER, NEC and EL tasks.
  \paragraph*{Nested entities.} The annotators were instructed not to annotate nested named entities. In fact, in cases of named entities such as \textit{Account of the Oxford Commemoration Concerts}, the only valid named entity would be the main one (in our example, the entity of type \textsc{publication} \textit{Account of the Oxford Commemoration Concerts}). Therefore, in our example, ``Oxford'' would not be annotated as an entity of type \textsc{city}.
  \paragraph*{Noisy entities.} The annotator is requested to record whether the named entity's superficial mention presents OCR errors. This strategy has allowed us to keep track of how many named entities' superficial mentions were \textit{noisy}, namely impacted by OCR errors (see Section \ref{subsub:stats_filtered}).
  \paragraph*{Named entity classification.} The named entities' types were assigned according to the AMR annotation guidance instructions and selected from the AMR named entity types list\footnote{Available at the URL \url{https://github.com/amrisi/amr-guidelines/blob/master/amr.md\#named-entities}, last time accessed on August 12th 2024.}. Sticking to the AMR annotation guidance instructions, the annotators were instructed to select the most specific type possible.
  \paragraph*{Entity linking.} The annotators were instructed to link the named entities recognized to their corresponding QID.
  \paragraph*{\texttt{NIL} links.} The annotators were instructed to assign the string \texttt{NIL} to named entities not present in Wikidata.

\section{\benchmarkEL{} Format}

\benchmarkEL{} is released in CoNLL-U format \footnote{\url{https://universaldependencies.org/format.html}} through its dedicated GitHub repository\footnote{Available at the URL \url{https://github.com/polifonia-project/historical-entity-linking/blob/main/benchmark/v1.0/mhercl_v1.0.tsv}.} and in JSONL format in HuggingFace\footref{foot:huggingface}.

\benchmarkEL{} CoNLL-U format release complies with HIPE-2022 data (based on IOB and CoNLL-U), facilitating future integration. An example of an annotated sentence of \benchmarkEL{} in CoNLL-U format is reported below:

\begin{quote}
\begin{lstlisting}[language=ConLL]

#document_id:DwightSJournalOfMusic__1873-016.txt_762
#document_date:1873
#sent_text:A native of Parma, ateighteen years of age Jong was received into the Conservatory of Music of that town, where Jong soon made himself.a name as the most promising pupil of the institution.
A	O	_
native	O	_
of	O	_
Parma	B-city	Q2683
,	O	_
ateighteen	O	_
years	O	_
of	O	_
age	O	_
Jong	B-person	NIL
was	O	_
received	O	_
into	O	_
the	O	_
Conservatory	B-school	Q1439627
of	I-school	Q1439627
Music	I-school	Q1439627
of	O	_
that	O	_
town	O	_
,	O	_
where	O	_
Jong	B-person	NIL
soon	O	_
made	O	_
himself.a	O	_
name	O	_
as	O	_
the	O	_
most	O	_
promising	O	_
pupil	O	_
of	O	_
the	O	_
institution	O	_
.	O	_
\end{lstlisting}
\end{quote}

In the above example, we can see that rows beginning with \textsc{\#} contain metadata. In more detail: 
\begin{itemize}
\item \textsc{\#document\_id}, provides the identifier (ID) of the document the sentence is extrapolated from;
\item \textsc{\#document\_date}, provides the date in which the document was published;
\item \textsc{\#sent\_text}, provides the text of the annotated sentence.\end{itemize}. 
The columns can be explained as follows:
\begin{itemize}
\item \textsc{Column 0} reports the sentence's tokens, one per row;
\item \textsc{Column 1}, reports IOB\footnote{According to IOB notation tagging format, each token which constitutes the beginning of a named entity string is assigned the label 'B' (an abbreviation of 'beginning'). Each token within a named entity string is labelled 'I' (an abbreviation for 'inside'). Each token outside a named entity string is labelled 'O' (an abbreviation for 'outside').} tags and type label of the named entity;
\item \textsc{Column 2}, reports the QID assigned by the annotators to the identified named entity.
\end{itemize}

\section{Named entities classification}
\label{app:A}

\subsection{Named entities types}

\begin{table}[h]
\centering
\caption{Named entity types occurring in \benchmarkEL{}}
\label{tab:ne_types_all}
\begin{tabularx}{\textwidth}{lX|lX}
\toprule
\textbf{Named Entity Type} & \textbf{\# Occurrences} & \textbf{Named Entity Type} & \textbf{\# Occurrences} \\
\midrule
person & 1246 & song & 3 \\
city & 251 & concert & 3 \\
music & 188 & location & 3 \\
organization & 94 & river & 3 \\
work-of-art & 85 & museum & 3 \\
country & 78 & newspaper & 3 \\
building & 52 & country-region & 3 \\
opera & 52 & symphony & 2 \\
theatre & 41 & religious-group & 2 \\
worship-place & 41 & thing & 2 \\
publication & 26 & person & 2 \\
book & 24 & family & 2 \\
road & 24 & language & 1 \\
company & 16 & band & 1 \\
school & 16 & province & 1 \\
city-district & 13 & island & 1 \\
magazine & 9 & park & 1 \\
event & 8 & empire & 1 \\
festival & 8 & hotel & 1 \\
street & 7 & scholarship & 1 \\
mountain & 6 & institution & 1 \\
university & 6 & village & 1 \\
government-organization & 5 & town & 1 \\
college & 4 & books & 1 \\
facility & 4 & person (fictional character) & 1 \\
local-region & 4 & lake & 1 \\
county & 4 & hall & 1 \\
continent & 4 & society & 1 \\
journal & 3 & military & 1 \\
square & 3 &  \\
\bottomrule
\end{tabularx}

\end{table}

The types assigned to named entities in \benchmarkEL{} are selected from the AMR named entity types list\footnote{We refer to their list as documented in the official GitHub repository, accessible at the URL \url{https://github.com/amrisi/amr-guidelines/blob/master/amr.md\#named-entities}}.
In Table \ref{tab:ne_types_all}, we report all the entity types assigned to named entities in \benchmarkEL{}.

\subsection{Named entities types taxonomy}

\begin{table}[ht]
    \centering
    \caption{Complete taxonomy used for expanding named entities' types to apply type constraints.}
    \begin{tabularx}{\textwidth}{lX}
    \toprule
    Type & Sub-types \\ \toprule
    B-event & B-concert, B-festival \\
    B-facility & B-street, B-road, B-park \\
    B-building & B-theatre, B-university, B-worship-place, B-museum, B-college, B-company, B-school, B-hall \\
    B-language & \\
    B-organization & B-theatre, B-university, B-worship-place, B-museum, B-college, B-company, B-school, B-empire, B-government-organization, B-religious-group, B-band \\
    B-person & \\
    B-publication & B-book, B-magazine, B-newspaper, B-journal \\
    B-work-of-art & B-music, B-opera, B-symphony, B-book, B-song \\
    B-location & B-park, B-hall, B-city, B-city-district, B-continent, B-country, B-county, B-local-region, B-mountain, B-road, B-square, B-country-region, B-province, B-island \\
    \bottomrule
    \end{tabularx}

    \label{tab:amr-taxonomy-complete}
\end{table}

As described in Section \ref{sec:type-c} of this paper, the named entity types are used in \modelCBLINK{} and \modelEL{} functionalities that assess the plausibility of the candidates according to their types. To optimize such a filter, we developed a taxonomy categorizing each named entity type occurring in \benchmarkEL{} into a hierarchy of types. When applying the type filter functionality, we expanded the dataset's gold type for each entity by including their more generic types from this taxonomy. The taxonomy is reported in Table \ref{tab:amr-taxonomy-complete}.

\section{Time-related Wikidata properties retrieval}

\begin{table}[h]
\caption{Time-related Wikidata properties used for candidate retention functionalities}
\label{tab:timerelated_wd_prop}
\begin{tabular*}{\textwidth}{@{\extracolsep\fill}lllp{5cm}}
\vtop{\hbox{\strut Retrieval}\hbox{Order}} & \vtop{\hbox{\strut Wikidata}\hbox{property}} & Name & Description\\
\toprule
1 & \textsc{P569} & \textsc{date of birth} & date on which the subject was born\\
2 & \textsc{P571} & \textsc{inception} & time when an entity begins to exist\\
3 & \textsc{P1619} & \textsc{date of official opening} & date or point in time an event, museum, theater etc. officially opened\\
4 & \textsc{P1191} & \textsc{date of first performance} & date a work was first debuted, performed or live-broadcasted\\
5 & \textsc{P10135} & \textsc{recording date} & the date when a recording was made\\
6 & \textsc{P577} & \textsc{publication date} & date or point in time when a work was first published or released\\
7 & \textsc{P575} & \textsc{time of discovery or invention} & date or point in time when the item was discovered or invented\\
8 & \textsc{P1317} & \textsc{floruit} & date when the person was known to be active or alive, when birth or death not documented\\
9 & \textsc{P7124} & \textsc{date of the first one} & qualifier: when the first element of a quantity appeared/took place\\
10 & \textsc{P10673} & \textsc{debut date} & date when a person or group is considered to have ``debuted"\\
11 & \textsc{P9448} & \textsc{introduced on} & date when a bill was introduced into a legislature\\
12 & \textsc{P6949} & \textsc{announcement date} & time of the first public presentation of a subject by the creator, of information by the media\\
13 & \textsc{P729} & \textsc{service entry} & date or point in time on which a piece or class of equipment entered operational service \\
14 & \textsc{P2031} & \textsc{work period (start)} & start of period during which a person or group flourished (fl. = ``floruit") in their professional activity\\
15 & \textsc{P585} & \textsc{point in time} & time and date something took place, existed or a statement was true\\
\bottomrule
\end{tabular*}
\end{table}

As described in Section \ref{sec:time-c} of this paper, time-related Wikidata properties are used in \modelCBLINK{} and \modelEL{} functionalities that assess the plausibility of the linking candidates according to their chronological consistency (time plausibility).

We report the time-related Wikidata properties that we retrieve in Table \ref{tab:timerelated_wd_prop}. The retrieval order is based on a qualitative analysis of the properties' names and descriptions to prioritize those values that can more precisely determine the plausibility of an entity, such as \textsc{P569 - date of birth} (ranked first), over more generic ones, like \textsc{P585 - Point in time} (ranked last).

\textcolor{orange}{\section{Prompt for LLMs evaluation on HEL}}
\label{app:prompt}

\textcolor{orange}{This appendix provides the prompt used to test LLMs in the experimental setting reported in \ref{sec:experiments}. The prompt, designed to query the model for the Wikipedia page name corresponding to a specific entity mention within a given sentence, is reported below:}

\textcolor{orange}{\noindent\texttt{In the sentence: <sentence> what is the Wikipedia page name of the entity <mention>? Answer with the page title only. If no Wikipedia page is appropriate, answer "NIL". Do not write anything else.}}

\end{appendices}

\end{document}